\documentclass{article}

     \PassOptionsToPackage{numbers, compress}{natbib}

\usepackage[preprint]{neurips_2025}

\usepackage[utf8]{inputenc} 
\usepackage[T1]{fontenc}    
\usepackage{hyperref}       
\usepackage{url}            
\usepackage{booktabs}       
\usepackage{amsfonts}       
\usepackage{nicefrac}       
\usepackage{microtype}      
\usepackage{xcolor}         
\usepackage[dvipsnames]{xcolor}

\usepackage{microtype}
\usepackage{graphicx}
\usepackage{subcaption}
\usepackage{multirow}
\usepackage{float}
\usepackage{bbm}
\usepackage{arydshln} 
\usepackage{multirow}
\usepackage{wrapfig,lipsum}
\usepackage{enumitem}
\usepackage{wrapfig}
\usepackage{pifont}

\usepackage{amsmath}
\usepackage{amssymb}
\usepackage{mathtools}
\usepackage{amsthm}
\usepackage{float}

\theoremstyle{plain}
\newtheorem{theorem}{Theorem}[section]

\newtheorem{lemma}[theorem]{Lemma}
\newtheorem{corollary}[theorem]{Corollary}
\theoremstyle{definition}
\newtheorem{definition}[theorem]{Definition}
\newtheorem{assumption}[theorem]{Assumption}
\theoremstyle{remark}
\newtheorem{remark}[theorem]{Remark}

\usepackage{algorithm}
\usepackage{algorithmic}
\usepackage{float}

\newcommand\independent{\protect\mathpalette{\protect\independenT}{\perp}}
\def\independenT#1#2{\mathrel{\rlap{$#1#2$}\mkern2mu{#1#2}}}
\DeclarePairedDelimiterX{\infdivx}[2]{(}{)}{%
  #1\;\delimsize\|\;#2%
}

\DeclarePairedDelimiter{\norm}{\lVert}{\rVert}

\usepackage{tikz}
\usetikzlibrary{shapes,decorations,arrows,calc,arrows.meta,fit,positioning}
\tikzset{
    -Latex,auto,node distance =1 cm and 1 cm,semithick,
    state/.style ={ellipse, draw, minimum width = 0.7 cm},
    point/.style = {circle, draw, inner sep=0.04cm,fill,node contents={}},
    bidirected/.style={Latex-Latex,dashed},
    el/.style = {inner sep=2pt, align=left, sloped}
}
\usetikzlibrary{decorations.pathreplacing}

\title{Synergizing Deconfounding and Temporal Generalization For Time-series Counterfactual Outcome Estimation}
\author{Yiling Liu \\
        Duke University\\
        Durham, NC, USA \\
        \texttt{yiling.liu@duke.edu} \\
      \And
      Juncheng Dong \\
      Duke University \\
      Durham, NC, USA \\
      \texttt{juncheng.dong@duke.edu}
      \And
      Chen Fu \\
      AI at Meta \\
      Sunnyvale, CA, USA \\
      \texttt{chenfu@meta.com}\\
      \And 
      Wei Shi \\
      AI at Meta \\
      Sunnyvale, CA, USA \\
      \texttt{weishi0079@meta.com}\\
      \And 
      Ziyang Jiang \\
      AI at Meta \\
      Menlo Park, CA, USA \\
      \texttt{jzy95310@meta.com}\\
      \And 
      Zhigang Hua \\
      AI at Meta \\
      Sunnyvale, CA, USA \\
      \texttt{zhua@meta.com}\\
       \And 
      David Carlson \\
      Duke University \\
      Durham, NC, USA \\
      \texttt{david.carlson@duke.edu}}
      
\begin{document}

\maketitle

\begin{abstract}
Estimating \emph{counterfactual} outcomes from time‑series observations is crucial for effective decision-making, e.g.\ when to administer a life‑saving treatment, yet remains significantly challenging because (\emph{i}) the counterfactual trajectory is never observed and (\emph{ii}) confounders evolve with time and  distort estimation at every step. To address these challenges, we propose a novel framework that \textbf{\emph{synergistically integrates}} two complementary approaches: \textbf{\emph{Sub-treatment Group Alignment}} (SGA)  and \textbf{\emph{Random Temporal Masking}} (RTM). 
Instead of the coarse practice of aligning marginal distributions of the treatments in latent space, SGA uses iterative \emph{treatment‑agnostic} clustering to identify fine-grained \emph{sub‑treatment groups}. Aligning these fine‑grained groups achieves improved distributional matching, thus leading to more effective deconfounding. We theoretically demonstrate that SGA optimizes a tighter upper bound on counterfactual risk and empirically verify its deconfounding efficacy. RTM promotes temporal generalization by randomly replaces input covariates with Gaussian noises during training. This encourages the model to rely less on potentially noisy or spuriously correlated covariates at the current step and more on stable historical patterns, thereby improving its ability to generalize across time and better preserve underlying \emph{causal relationships}. Our experiments demonstrate that while applying SGA and RTM individually improves counterfactual outcome estimation, their synergistic combination consistently achieves state-of-the-art performance. This success comes from their distinct yet complementary roles: RTM enhances temporal generalization and robustness across time steps, while SGA improves deconfounding at each specific time point.
\end{abstract}



\section{Introduction}\label{sec:intro}

Estimating causal effects from time series data is a pivotal task in many fields such as healthcare, politics, and economics \citep{bisgaard2011time, freeman1983granger, morid2023time}. This fundamentally requires the ability to estimate counterfactual outcomes: what would have happened under different interventions over time. For example, in the treatment of \emph{Ductal Carcinoma In Situ}, accurately estimating counterfactual outcomes is crucial for determining the timing of surgical intervention: if surgery is too late, the cancer may progress to an invasive stage; if conducted too early, the procedure may be unnecessarily invasive \citep{grimm2022ductal}. 

Motivated by this, we explore counterfactual outcome estimation in time series from observational data. The success of causal inference in time series relies on \textbf{effective reduction of time-dependent confounding}. However, this task is challenging, primarily because of the unobservability of counterfactual outcomes and the time-varying confounding in time series. A well-established approach for reducing confounding in \emph{static} causal inference is to minimize an upper bound on the counterfactual estimation error~\citep{johansson2016learning, johansson2022generalization, li2017matching, yao2018representation}, which can be decomposed into two key components: ($i$)  the \emph{factual loss} and ($ii$)  the \emph{statistical discrepancy between treatment and control groups} in the learned representation space. Algorithmically, these methods simultaneously ($i$) minimize the prediction error of the factual outcomes and ($ii$) align the treatment groups in the latent space. By ensuring that the representations of multiple treatment groups are brought closer together, they provably reduce the bias introduced by confounders~\citep{johansson2022generalization}. Building on this idea to reduce confounding for time series, existing approaches aim to learn representations that remain invariant to the treatment assignment \textbf{at each time step}~\citep{bica2020estimating, melnychuk2022causal}. However, in practice, they typically result in optimizing relatively \textbf{loose} upper bounds on the counterfactual error at individual time steps \citep{arjovsky2017towards} with adversarial training. 
Moreover, models may \textbf{over-rely on contemporaneous information} at each time step, potentially hindering their ability to learn stable long-term patterns across time points \citep{ghouse2024understanding}. This can lead to \textbf{compromised generalization} for estimating long-term effects, as errors can propagate and compound over time.

\begin{figure*}[t]
    \centering
    \includegraphics[width=0.78\textwidth]{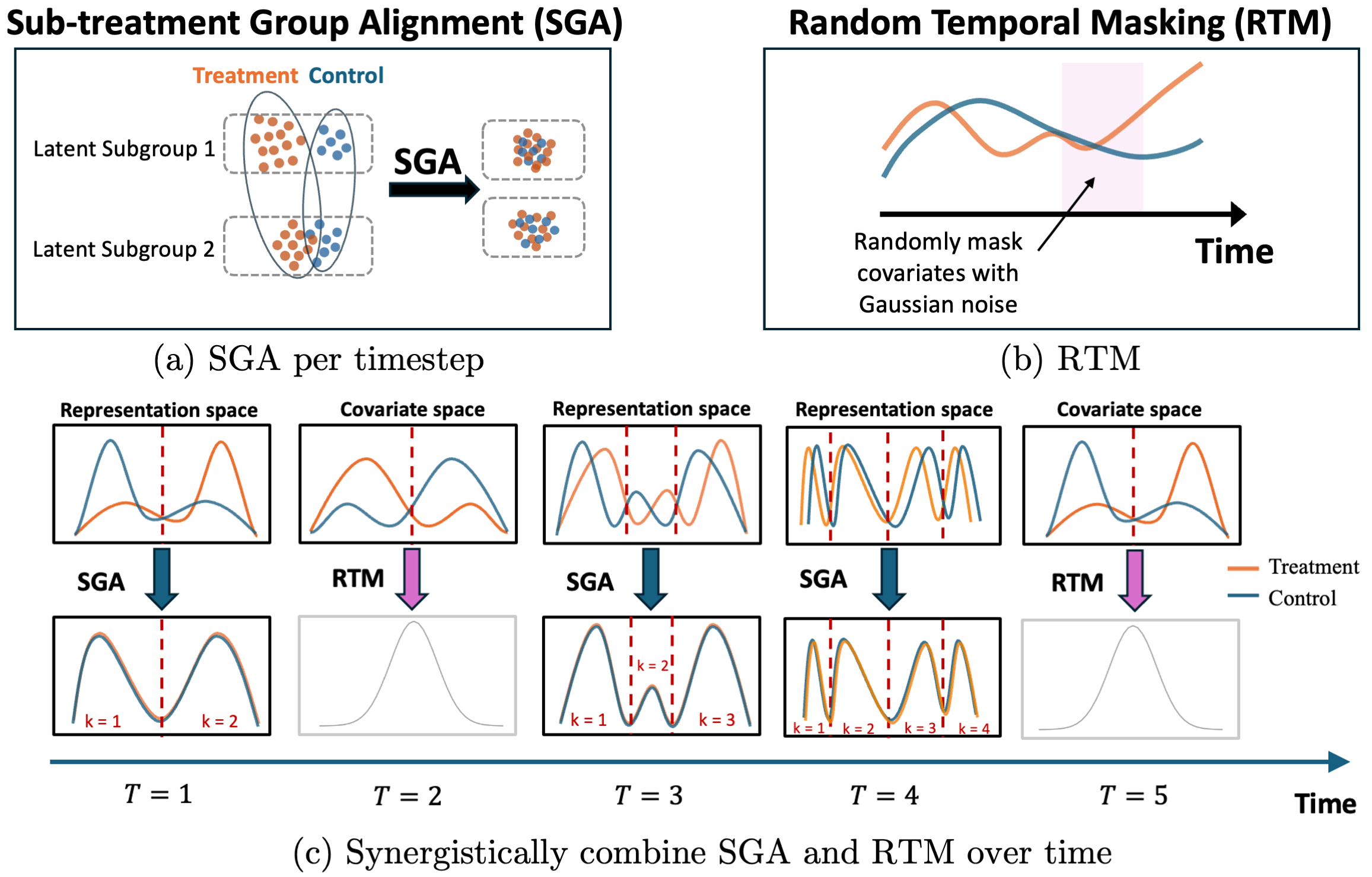}
    \caption{\textbf{Conceptual overview of SGA and RTM.} {\bf (a)} SGA identifies and aligns fine-grained sub-treatment groups at each timestep to improve deconfounding. {\bf (b)} RTM forces the model to leverage historical patterns and enhancing temporal generalization. {\bf (c)} SGA and RTM are synergistically combined to improve counterfactual outcome estimation. Here, $k$ denotes sub-treatment group index.}
    \vspace{-5mm}
\label{fig1}
\end{figure*}

To address these challenges, we introduce our framework with two novel approaches:
\begin{itemize}[left=5pt,topsep=0.1em,parsep=0pt]
\item[\ding{118}] \textbf{\emph{Sub-treatment Group Alignment (SGA)}}, which improves deconfounding at each individual time point by identifying and subsequently aligning \emph{sub-treatment groups}.
\item[\ding{79}] \textbf{\emph{Random Temporal Masking (RTM)}} promotes temporal generalization and robust learning from historical patterns by randomly masking covariates at selected time points with Gaussian noise.
\end{itemize}


\textbf{Sub-treatment Group Alignment (SGA).} SGA first identifies \textbf{sub-treatment groups in the representation space} through \emph{treatment‑agnostic} clustering at each timestep, and then subsequently aligning the corresponding sub-groups across different treatment groups (see Figure~\ref{fig1}). In Section~\ref{sec:bounds}, we establish that such sub-group alignment indeed leads to a tighter bound on the counterfactual estimation error. This more fine-grained alignment enables \textbf{improved matching} of treatment groups, thus allowing us to \textbf{reduce the estimation error} more effectively than existing methods. 

\textbf{Random Temporal Masking (RTM).}
While SGA addresses confounding at \textbf{individual} time points, RTM enhances the model's ability to generalize \textbf{across} time series. Inspired by masked language modeling, RTM uses random covariate masking, where input covariates at randomly selected time points are replaced with Gaussian noise during training. 
There are multiple perspectives to understand the benefits of RTM: ($i$) When input covariates at certain time points are masked, the models must extract useful information from past observations to predict future factual outcomes. In other words, we encourage the model to \textbf{focus on the causal relationships that span across time}, leading to better counterfactual predictions. 
($ii$) RTM can prevent model from becoming overly reliant on information from current time points, thus \textbf{reducing overfitting to the factual distribution}. 


\textbf{Empirical Validation.} Our framework is \textbf{broadly applicable} and designed for \textbf{straightforward integration} into existing representation learning-based frameworks for time series counterfactual estimation. We empirically validate our framework through comprehensive experiments on synthetic and semi-synthetic datasets, and it consistently demonstrates state-of-the-art (SOTA) performance.

\textbf{Organization.} We first formally define the problem in Section~\ref{sec:problem-setup} and review related works in Section~\ref{sec:related_work}. Then in Section~\ref{sec:bounds}, we theoretically establish how sub-treatment group alignment achieves improved deconfounding, thus motivating our SGA approach. In Section~\ref{sec:framework}, we present our framework with SGA and RTM. Finally, experimental results in Section~\ref{sec:experiments} show that applying SGA and RTM individually enhances performance, and that their \textbf{\emph{\color{black}synergistic combination}} achieves SOTA results.

\section{Problem Setup}\label{sec:problem-setup}

\textbf{Notations.} We use upper-case letters (e.g., $A, Y$) for scalar random variables and lower-case letters (e.g., $a, y$) for their corresponding realizations. Multi-dimensional random variables and realizations are denoted using bold fonts (e.g., $\boldsymbol{X}$ and $\boldsymbol{x}$). 

\textbf{Observational Data.} We consider a dataset containing $N$ samples following conventional setups \citep{bica2020estimating, li2020g, melnychuk2022causal}. Observations are recorded over $T$ time steps, i.e., $t = 1,...,T$. At each time $t$, a discrete treatment $A_{t} \in \mathcal{A} = \{a_0, a_1,...,a_{|\mathcal{A|}-1}\}$ is assigned. Thus, for each $i$, we observe time-varying covariates $\mathbf{X}_{t}^{(i)} \in \mathbb{R}^d $, factual treatment $A_{t}^{(i)}$, and outcome $Y_{t}^{(i)}$ of the factual treatment.

We use the following notation to represent the history up to time step \( t \) for each unit \( i \),
\begin{equation*}
\textstyle
\bar{\mathbf{H}}_{t}^{(i)} = \{ \bar{\mathbf{X}}_{t}^{(i)}, \bar{\mathbf{Y}}_{t}^{(i)}, \bar{\mathbf{A}}_{t-1}^{(i)}, \mathbf{V}^{(i)} \}
\end{equation*}
    where \( \bar{\mathbf{X}}_{t}^{(i)} = \{ \mathbf{X}_{s}^{(i)} : s \leq t \} \) denotes the sequence of time-varying covariates up to time \( t \),
    \( \bar{\mathbf{Y}}_{t}^{(i)} = \{ Y_{s}^{(i)} : s \leq t \} \) represents the sequence of observed outcomes up to time \( t \),
    \( \bar{\mathbf{A}}_{t-1}^{(i)} = \{ A_{s}^{(i)} : s \leq t - 1 \} \) is the sequence of treatments up to time \( t - 1 \),
    \( \mathbf{V}^{(i)} \in \mathbb{R}^p \) denotes the static covariates.



\textbf{Objective.} Given the history up to current time \( t \) and assuming a specific treatment sequence \( \mathbf{a}_{t:t+\tau - 1}^{(i)} \) from time \( t \) to \( t + \tau - 1 \) applied to sample $i$, \textbf{\color{black}our goal is to estimate}, for each unit \( i \), \textbf{\color{black}the future outcome at time step $t+\tau$}. That is, \( \tau \) time steps after the current time \( t \). To ensure that these counterfactual outcomes are identifiable, we follow the \emph{potential outcomes framework} and make several standard assumptions to support identifiability~\citep{rosenbaum1983central, rubin2005causal}. Due to space constraint, details on the assumptions are provided in Appendix~\ref{sec:potential_outcome}. Specifically, we estimate
\begin{equation}
\textstyle
\mathbb{E} [ Y_{t + \tau}^{(i)} ( \mathbf{a}_{t:t+\tau - 1}^{(i)}) \,|\, \bar{\mathbf{H}}_{t}^{(i)} ], 
\end{equation}
where \( Y_{t + \tau}^{(i)} ( \mathbf{a}_{t:t+\tau - 1}^{(i)} ) \) is the potential outcome at \( t + \tau \) for unit \( i \) with treatment sequence \( \mathbf{a}_{t:t+\tau - 1}^{(i)} \).

\section{Related Work}
\label{sec:related_work}
We review the \textbf{most} relevant work below and provide a \textbf{comprehensive} discussion in Appendix \ref{sec:complete_related_work}.


\textbf{Estimating Counterfactual Outcomes Over Time.} Estimating counterfactual outcomes in time-series is challenging due to time-varying confounders. Traditional methods such as G-computation and marginal structural models \citep{hernan2001marginal, robins1986new, robins2008estimation, robins2000marginal,xu2016bayesian} often lack flexibility for complex datasets and rely on strong assumptions. To address these limitations, researchers have developed models that build on the potential outcomes framework initially proposed by \citet{rubin1978bayesian} and extended to time series by \citet{robins2008estimation}. Notable among recent methods are Recurrent Marginal Structural Networks (RMSNs) \citep{lim2018forecasting}, G-Net \citep{li2020g}, Counterfactual Recurrent Networks (CRN) \citep{bica2020estimating}, and the Causal Transformer (CT) \citep{melnychuk2022causal}, which use approaches such as propensity networks and adversarial learning to mitigate the effects of time-varying confounding. However, practical challenges with adversarial training can affect the stability of causal effect estimations. Specifically, training adversarial networks can be challenging due to issues such as mode collapse and oscillations \citep{liang2018generative}. Additionally, adversarial training minimizes the Jensen-Shannon divergence (JSD) only when the discriminator is optimal \citep{arjovsky2017towards}, which may not always be achievable in practice; even when the discrminator is optimal, using JSD optimizing relatively loose upper bounds on the counterfactual error. To address these challenges, we propose using the Wasserstein-1 distance and provides stronger theoretical guarantees \citep{mansour2012multiple, redko2017theoretical}. 

\textbf{Masked Language Modeling.} Masked language modeling (MLM) is a common self-supervised pre-training technique for large language models. It operates by randomly masking certain words or tokens in the input, with the model trained to predict the masked tokens. BERT \citep{devlin2018bert} is the most well-known model that uses this technique. Recent studies have also demonstrated the effectiveness of MLM in enhancing generalization across sequence-based tasks. For example, \citet{chaudhary2020dict} shows that when combined with cross-lingual dictionaries, MLM improves predictions for the original masked word and also generalizes to its cross-lingual synonyms. Inspired by the success of masking strategies in language models, we introduce Random Temporal Masking (RTM) for time-series data. Unlike MLM, which focuses on predicting the masked inputs, RTM encourages the model to focus on information that is crucial for both the current time point and future time points, preserve causal information, and reduce the risk of overfitting to factual outcomes.

\section{Theoretical Motivation for Sub-treatment Group Alignment}\label{sec:bounds}

In this section, we provide a theoretical motivation for our proposed Sub-treatment Group Alignment (SGA) method, rigorously illustrating that \textbf{aligning sub-treatment groups in the latent space leads to more effective deconfounding in estimating counterfactual outcomes} over time series. 

\textbf{From Static to Time Series.} 
By aligning the corresponding sub-treatment groups, \textbf{SGA achieves improved alignment} and thus more effective deconfounding. Since existing time series methods often apply alignment independently at each time step \citep{bica2020estimating, melnychuk2022causal}, demonstrating SGA's superiority in a static context provides a strong foundation for its benefits in time-series. 
In other words, achieving better deconfounding at \emph{individual time steps} will consequently lead to more effective deconfounding \emph{over the entire time series}.


\textbf{Section Organization.} 
Given the rationale that time‑series alignment can be decomposed into a collection of static sub‑problems, \textbf{it is sufficient to consider static settings}. Thus, in Section~\ref{sec:static-setting} we briefly review ($i$) representation learning-based models that use alignment ($ii$) why alignment mitigates confounder bias in the static setting. Subsequently, in Section~\ref{sec:subgroup}, we theoretically establish that SGA indeed improves alignment in static settings. This implies that \textbf{integrating SGA into existing time-series frameworks can improve deconfounding at each step}, leading to overall improvements.


\subsection{Alignment for Static Setting}\label{sec:static-setting} 
Since there is only one time step $t=1$ in the static setting, we will omit all notations about the time step for clarity. We will use the Wasserstein-1 distance $W_1$ to measure the statistical discrepancy between two random variables. Due to space constraints, we defer the mathematical definition of $W_1$ to Appendix~\ref{sup_def}, and a discussion of SGA's computational cost to Appendix~\ref{sec:cost}.

\textbf{Representation Learning-based Models.} Let $\Phi:\mathcal{X} \rightarrow \mathcal{R}$ be a representation-learning function and $h : \mathcal{R} \times \{0, 1\} \rightarrow Y$ be an hypothesis. We have $h(\Phi(x),a)$ as a predictor for an individual $x$'s potential outcome under treatment assignment $a$. The goal is to find a pair of  $(h,\Phi)$ that optimizes both the \emph{factual loss} $\epsilon_{F}^{\star}(h, \Phi)$ and \emph{counterfactual loss} $\epsilon_{CF}(h, \Phi)$, which are defined in Appendix~\ref{def:f_cf} and~\ref{def:new_error_f}. Note that low factual and counterfactual losses are both necessary and sufficient conditions for accurate potential outcome prediction \citep{aloui2023transfer}.

\textbf{Counterfactual Error Estimation.} However, the \textbf{counterfactual loss $\boldsymbol{\epsilon_{CF}(h, \Phi)}$ cannot be directly optimized} because the counterfactual outcomes are not observable in real-world scenarios. To this end, a group of well-established approaches minimize \textbf{upper bounds} of $\epsilon_{CF}(h, \Phi)$. These approaches are mainly based on the following result from~\citet{shalit2017estimating}, restated here. 

\begin{theorem}[Simplified Lemma A8 from~\citet{shalit2017estimating}, complete version provided in Appendix~\ref{thm:complete_theorem_1}.]
\label{thm:theorem1}
Let $\Phi:\mathcal{X} \rightarrow \mathcal{R}$ be a one-to-one and Jacobian-normalized representation function. Let $h : R \times \{0, 1\} \rightarrow Y$ be a hypothesis with Lipschitz constant:
\begin{equation}
\label{eqn:original_bound}
\epsilon_{CF}(h, \Phi) \leq \epsilon_{F}^{\star}(h, \Phi) + 2 \cdot B_\Phi \cdot W_1(p_\Phi^{0}, p_\Phi^{1}),
\end{equation}
where $B_\Phi$ is a constant and $p_{\Phi}^{a}$ is the distribution of the random variable $\Phi(X)$ conditioned on $A=a$, that is, representations for individuals receiving treatment $a \in \{0,1\}$.
\end{theorem}
\textbf{Motivation for Alignment.} This theorem implies that a model $(\Phi,h)$ has low counterfactual error if \textbf{\emph{(a)}} it has \textbf{low factual error} (which can be easily achieved by minimizing the prediction error on the observational data) and \textbf{\emph{(b)}} the covariates of individuals from distinct treatment groups are \textbf{statistically similar to each other in the latent (representation) space}. Motivated by these, representation learning-based methods aim to align the treatment and control groups in the latent space while minimizing the factual error. In particular, successful alignment and low factual error guarantee a small value for the upper bound in Equation~(\ref{eqn:original_bound}), implying the model has low counterfactual error. However, in practice, \textbf{\emph{\color{black}the error bound may be loose}}, leaving the model performance suboptimal \citep{arjovsky2017towards}.

\subsection{Benefits of Sub-treatment Group Alignment}\label{sec:subgroup}

To this end, we propose to use the \textbf{sub-treatment group structures} to achieve \textbf{tighter} counterfactual error bound, thus supporting \textbf{more effective} alignment. 

\textbf{Sub-treatment Groups.} We hypothesize that each treatment group is a mixture of $K$ sub-treatment groups in the latent space, and that \textbf{the sub-treatment groups across different treatment groups correspond to one another}. For example, in medical studies, patients may naturally form sub-groups \textbf{before} the beginning of experiments, based on latent variables such as demographic characteristics or genetic factors. Consider a scenario where patients are sub-grouped according to age (e.g., children, adults, seniors), gender, or genetic markers that influence their response to treatment. Even though these patients receive different treatments, the underlying characteristics defining the sub-groups are consistent across treatment groups. By aligning these corresponding sub-groups in the latent space, we can more effectively account for \textbf{hidden confounders} like genetic predispositions or socio-demographic factors, leading to more accurate estimation of treatment effects.

Recall $p_{\Phi}^{a}$ is the distribution of representations for individuals receiving treatment $a \in \{0,1\}$, thus
\begin{equation*} 
\textstyle
p_{\Phi}^{0} = \sum_{k=1}^{K} w_k^{0} P_{\Phi,k}^{0}, \quad p_{\Phi}^{1} = \sum_{k=1}^{K} w_k^{1} P_{\Phi,k}^{1}, \end{equation*}
where for $a \in \{0,1\}$, $w_k^{a}$ represents the proportion of the $k$-th sub-group in treatment group $a$, and $P_{\Phi,k}^{a}$ is the distribution of representations of individuals in the $k$-th sub-group under treatment $a$.

\textbf{Sub-treatment Group Alignment (SGA).} SGA has the following alignment objective:
\begin{equation}\label{eqn:sga}
\textstyle \sum_{k=1}^K w_k^{1} W_1 ( P_{\Phi,k}^{0}, P_{\Phi,k}^{1} ). \end{equation}
In particular, SGA minimizes the \textbf{weighted sum} of the Wasserstein distances between these \textbf{corresponding sub-treatment groups}. By aligning on a sub-treatment group level, SGA achieves more refined alignment. Indeed, motivated by the generalization bound in the field of \emph{domain adaptation}~\citep{liu2025understanding}, we next prove in Theorem~\ref{thm:weight_subdomain_distance_stronger} that SGA is at least as tight as the original alignment under reasonable assumptions, thus resulting in more effective deconfounding.

\begin{theorem}[SGA Improves Generalization Bounds]
\label{thm:weight_subdomain_distance_stronger}
Under weak assumptions, the following inequalities hold:
\begin{equation*}
\begin{split}
&\epsilon_{CF}(h, \Phi) \le \epsilon_{F}(h, \Phi)  + 2 B_\Phi (\textstyle \sum_{k=1}^K w_k^{1} W_1(P_{\Phi,k}^{0}, P_{\Phi,k}^{1})) \\
& \textstyle\sum_{k=1}^K w_k^{1} W_1(P_{\Phi,k}^{0}, P_{\Phi,k}^{1}) \leq W_1(p_\Phi^{0}, p_\Phi^{1}) + \delta_c,
\end{split}
\end{equation*}
where $B_{\Phi}$ is the same constant in Theorem~\ref{thm:theorem1} and $\delta_c$ is $4\sqrt{\epsilon}$. 
\end{theorem}
See Appendix~\ref{sup_def} for proof of Theorem~\ref{thm:weight_subdomain_distance_stronger} and detailed assumption statements and interpretations.

\begin{remark}
Theorem~\ref{thm:weight_subdomain_distance_stronger} proves that SGA improves the original counterfactual error bound in Theorem~\ref{thm:theorem1} by optimizing an upper bound that is at least as tight as the original bound. In Appendix~\ref{emp_tighter_more}, we provide empirical evidence that SGA indeed results in a \textbf{much tighter} upper bound compared to the original counterfactual error bound. 
\end{remark}

\section{Framework}\label{sec:framework}

\begin{wrapfigure}{r}{0.55\textwidth}
\centering
\includegraphics[width=0.55\textwidth]{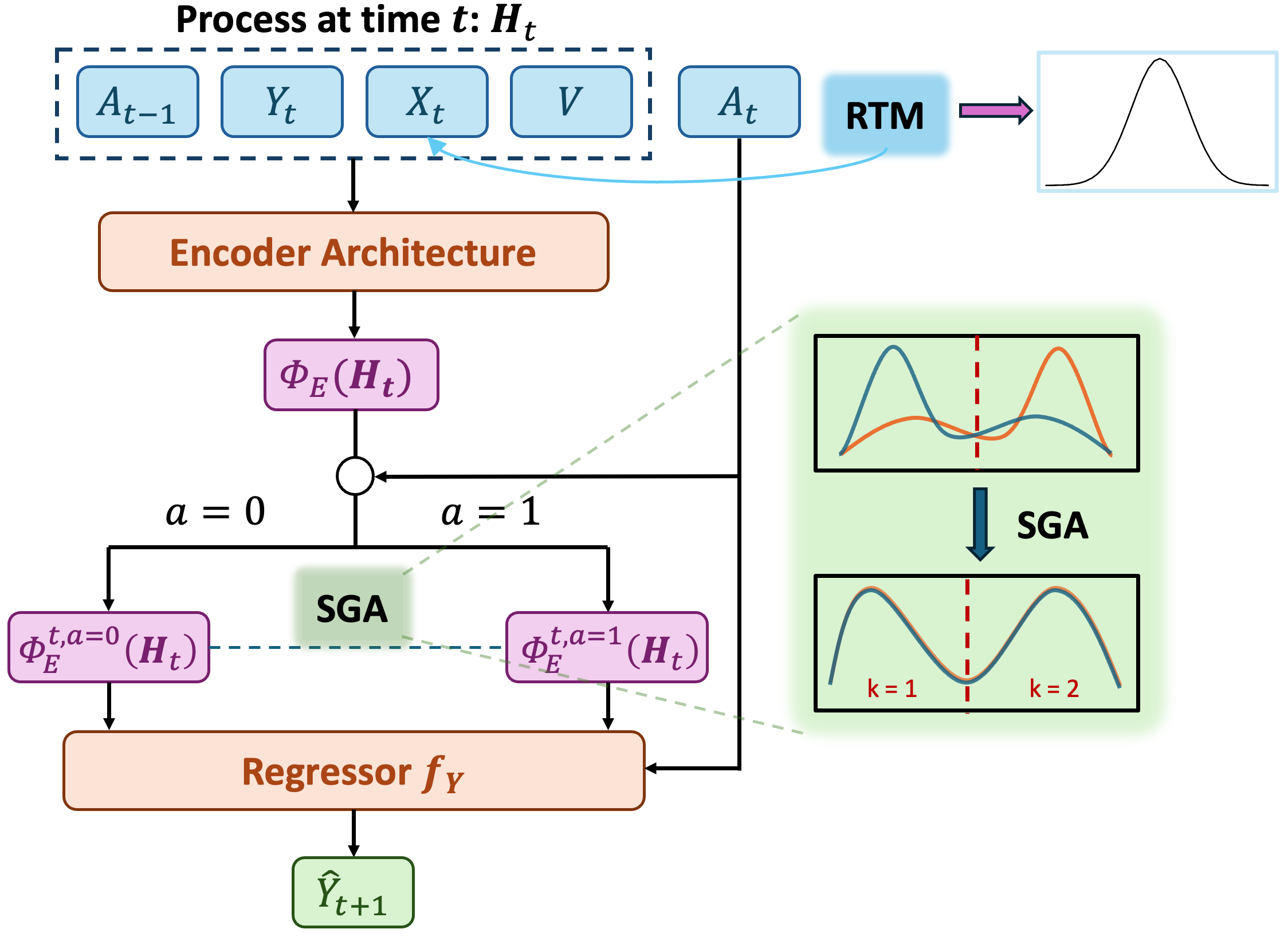}
\caption{\textbf{Overview of SGA \& RTM} at each timepoint. For simplicity, we show a binary treatment scenario.} 
\vspace{-3mm}
\label{fig:2}
\end{wrapfigure}

We propose a framework that \emph{\textbf{synergistically integrates}} our two novel approaches: Sub-treatment Group Alignment (SGA) and Random Temporal Masking (RTM). Figure~\ref{fig:2} illustrates the general architecture of our framework. 

\textbf{Model Architecture.} A key feature of our framework is its versatility; it is not restricted to a specific architecture and can be integrated with various representation-based methods for time series causal inference. 
Typically, such methods consist of a \emph{time series encoder} $\phi_E$ (parameterized by $\theta_E$) that learns representations from input time series data, and an \emph{outcome regressor} $f_Y$ (parameterized by $\theta_Y$) that predicts outcomes at the next time point. We note that the encoder $\phi_E$ can be instantiated with any sequence model architecture, such as RNNs, LSTMs \citep{hochreiter1997long}, or transformers \citep{vaswani2017attention}. In Section~\ref{sec:experiments}, we demonstrate this flexibility by integrating our approaches with two well-established for time series causal inference: \emph{Causal Transformer} ({\bf CT})~\citep{melnychuk2022causal} and \emph{Counterfactual Recurrent Networks} 
 ({\bf CRN})~\citep{bica2020estimating}.

\textbf{Objective Function.} At each time step $t$, our framework optimizes the objective
\[\textstyle
\min_{\theta_Y, \theta_E} L^{t}_Y(\theta_Y, \theta_E) + \lambda L^{t}_D(\theta_E),
\]
where $L^{t}_Y$ represents the \textbf{\color{black}factual outcome loss} and $L^{t}_D$ denotes the \textbf{\color{black}SGA loss} calculated with the Wasserstein-1 distance, balanced by $\lambda$. Details are provided below.

\textbf{Factual Outcome Loss.} At each time step $t$, the model learns to predict the observed outcomes, conditioned on $\mathbf{H}_t^{(i)}$ which contains the information from previous steps and the current covariates,  by optimizing 
\[\textstyle
L^{t}_Y(\theta_Y, \theta_E) = \frac{1}{N} \sum_{i=1}^{N} (\ell(y_i^{t+1}, \hat{y}_i^{t+1})),
\]
where $\hat{y}_i^{t+1} = f_Y\left(\phi_E\left(\mathbf{H}_t^{(i)}, A_t^{(i)}\right)\right)$ and $\ell(\cdot,\cdot)$ denotes the loss function (e.g., mean squared error).

\textbf{SGA Loss.} Motivated by Section~\ref{sec:bounds}, our framework aligns the sub-treatment groups across distinct treatment groups. To this end, at each time step $t$ and for each treatment group $a$, we use Gaussian Mixture Models (GMMs) to cluster the individuals' features in the representation space into $K$ sub-treatment groups. Let the random variable $\phi_E^{t,a,k}(\mathbf{H}_t)$ denote the representations of samples in the $k$-th sub-group of treatment group $a$ at time step $t$. 

To accommodate the applications with \textbf{\color{black}multiple treatment groups} (more than two), for each time step $t$ and each corresponding sub-treatment group, we align the sub-treatment groups with \textbf{the uniform mixtures of them}. That is, for all the $k$-th sub-treatment groups in all $|\mathcal{A}|$ treatment groups where $|\mathcal{A}|$ is the total number of treatments, we first create a mixture of them with uniform weights and align all of them with the uniform mixture. Note that by triangle inequality this is a \textbf{sufficient condition} to align multiple groups well. Specifically, the SGA loss is defined as:
\[\textstyle
L^{t}_D(\theta_E) = \sum_{k=1}^K\sum_{a \in \mathcal{A}}  w_k^{t,a} W_1(\phi_E^{t,a,k}(\mathbf{H}_t),\phi_E^{t,k}(\mathbf{H}_t)),
\]
where $w_k^{t,a}$ represents the proportion of samples in sub-group $k$ of treatment group $a$, and $\phi_E^{t,k}(\mathbf{H}_t)$ is the uniform mixture of $\{\phi_E^{t,a,k}(\mathbf{H}_t)\}_{a\in\mathcal{A}}$. Note that all the quantities in $L^{t}_D(\theta_E)$ can be estimated from the observational data. We provide implementation details and our algorithm in Appendix \ref{sec:algo}.

\section{Experiments}\label{sec:experiments}

\textbf{Main Results.} We evaluate our framework on a fully-synthetic Pharmacokinetic-Pharmacodynamic (PK-PD) benchmark \citep{bica2020estimating, geng2017prediction} and a semi-synthetic dataset derived from \textsc{MIMIC-III} \citep{johnson2016mimic}. We integrate our framework into the architectures of the LSTM-based \emph{Counterfactual Recurrent Networks (CRN)} \citep{bica2020estimating} and the Transformer-based \emph{Causal Transformer (CT)} \citep{melnychuk2022causal}. Across both datasets, our experiments consistently demonstrate that the \textbf{\emph{synergistic combination}} of SGA and RTM  achieves state-of-the-art (SOTA) performance in counterfactual outcome estimation (Section~\ref{sec:experiments_fully_syn} and \ref{sec:experiments_semi_syn}). Full details are deferred to Appendix~\ref{supp:experiments}.

\textbf{Analysis.} We run comprehensive ablation studies, showing {\bf (a)} individual contributions of SGA and RTM (Section~\ref{sec:ablation_studies_summary}), {\bf (b)} detailed analyses of RTM's masking strategy (Section~\ref{sec:masking_strategies_summary}), {\bf (c)} RTM's impact on attention mechanisms (Section~\ref{sec:rtm_attention_summary}), and {\bf (d)} SGA's sensitivity to clustering choices (Section~\ref{sec:clustering_sensitivity_summary}).


\textbf{Baseline Methods. } 
We compare against SOTA baselines: Marginal Structural Models ({\bf MSMs})~\citep{hernan2001marginal, robins2000marginal}, Recurrent Marginal Structural Networks ({\bf RMSNs}) \citep{lim2018forecasting}, G-Net \citep{li2020g}, Counterfactual Recurrent Networks ({\bf CRN}) \citep{bica2020estimating}, and Causal Transformer ({\bf CT}) \citep{melnychuk2022causal}.

\subsection{Experiments on Fully-Synthetic Data from a PK-PD Model of Tumor Growth}
\label{sec:experiments_fully_syn}
We first evaluate on the fully-synthetic data frequently used in the literature~\citep{bica2020estimating, melnychuk2022causal}, which allows simulation of treatment-response dynamics and varying levels of time-dependent confounding.



\textbf{Tasks and Evaluation Metrics}. Following \citet{melnychuk2022causal}, we report normalized Root Mean Squared Error (RMSE) on both \emph{one‑step-ahead} and \emph{$\tau$‑step‑ahead} counterfactual predictions under \emph{varying levels} of time-varying confounding, indexed by $\gamma$. Details on dataset generation, hyperparameters, visualization of the representation space, and results with \emph{error bars} are in Appendix~\ref{sup:exp_fully_synthetic}. 


\begin{wrapfigure}{hr}{0.53\textwidth}
    \centering
    \vspace{-6mm}
    \includegraphics[width=0.53\textwidth]{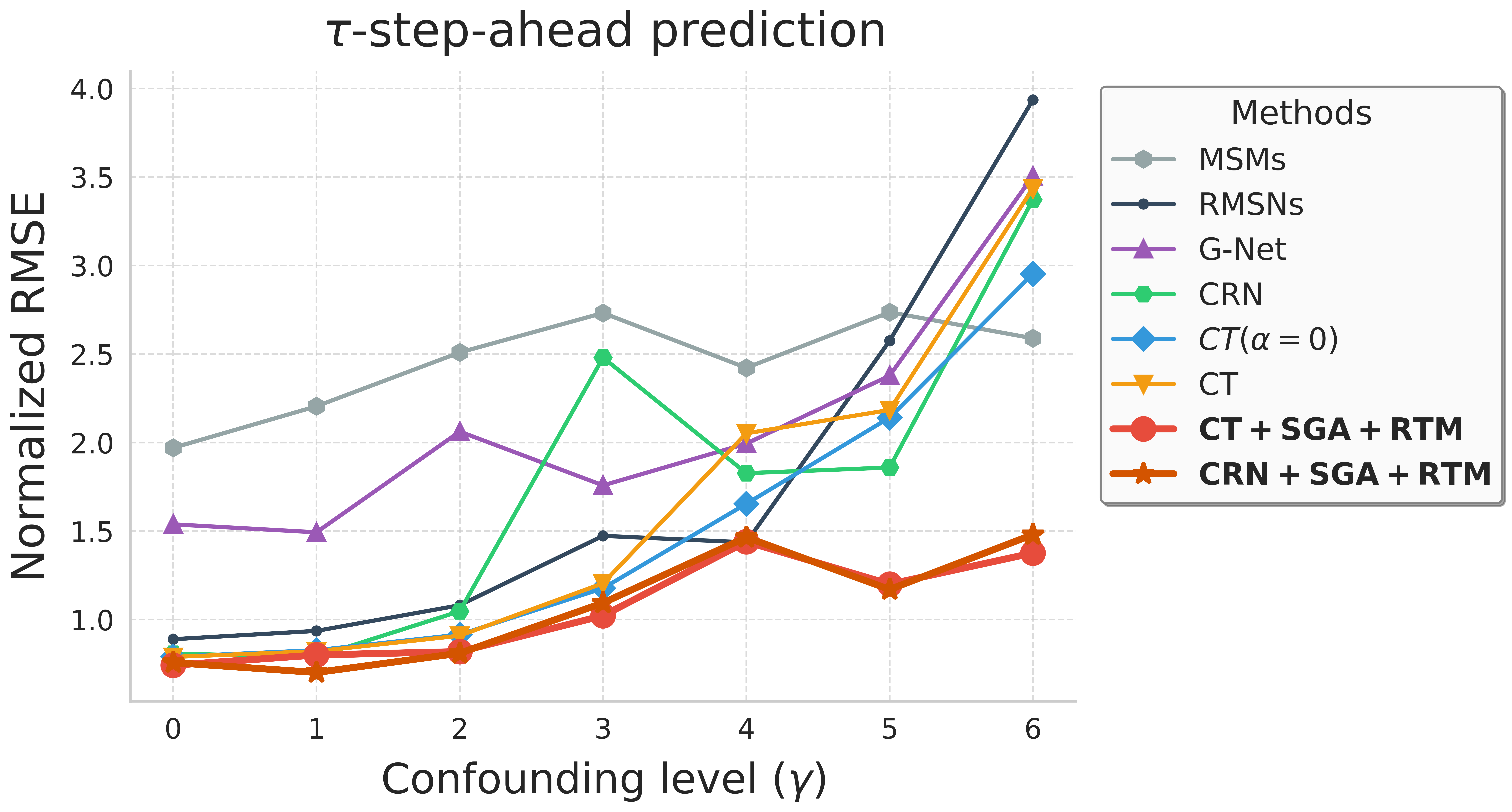}
    \vspace{-5mm} 
    \caption{Performances on $\tau$-step-ahead ($\tau$=6) prediction. Note that CT ($\alpha$=0) refers to CT w/o alignment.}
\label{fig:3}
\end{wrapfigure}

\textbf{Results.} As shown in Figure~\ref{fig:3} and Table \ref{tab:fully_synthetic_single_sliding_treatment}, integrating both SGA and RTM into CRN and CT \textbf{\color{black}significantly improves} 
their performance compared to the vanilla models. Notably, our framework performs exceptionally well in scenarios with \textbf{\color{black}high levels of confounding}, indicating our effectiveness in deconfounding. Baselines results for confounding levels $\gamma \in [0,4]$ are cited from \citet{melnychuk2022causal}, and we extend experiments to higher confounding levels to more thoroughly test the robustness of our methods in complex causal scenarios.



\begin{table}[ht]
\centering
\vspace{-2mm}
\caption{Normalized RMSE for one-step-ahead and $\tau$-step-ahead predictions on fully synthetic data, comparing vanilla CRN/CT with CRN/CT enhanced with SGA + RTM. The values in \textbf{\textcolor{blue}{blue}} indicate lower RMSE for CRN-based models, and values in \textbf{\textcolor{violet}{violet}} indicate lower RMSE for CT-based models. }
\vspace{1mm} 
\scalebox{0.72}{
\begin{tabular}{|c|c|c|c|c|c|c|c|c|}
\hline
$\boldsymbol{\tau}$ & \textbf{Method} & $\boldsymbol{\gamma = 0}$ & $\boldsymbol{\gamma = 1}$ & $\boldsymbol{\gamma = 2}$ & $\boldsymbol{\gamma = 3}$ & $\boldsymbol{\gamma = 4}$ & $\boldsymbol{\gamma = 5}$ & $\boldsymbol{\gamma = 6}$\\
\hline
\multirow{4}{*}{$\boldsymbol{\tau = 1}$} 
& \textbf{CRN}           &  0.755 & 0.788 & 0.881 & 1.062 &	1.358 & 1.403 & 2.031\\
& \textbf{CRN + SGA + RTM} & \textbf{\textcolor{blue}{0.720}} &  \textbf{\textcolor{blue}{0.741}} & \textbf{\textcolor{blue}{0.822}} & \textbf{\textcolor{blue}{0.802}}  & \textbf{\textcolor{blue}{0.949}} & \textbf{\textcolor{blue}{1.107}} & \textbf{\textcolor{blue}{1.808}}\\
\cdashline{2-9} 
& \textbf{CT}   & 0.770 & 0.783 & 0.864 & 1.098 & 1.413 & 1.497 & 2.204 \\
& \textbf{CT + SGA + RTM}  &  \textbf{\textcolor{violet}{0.750}} & \textbf{\textcolor{violet}{0.736}} & \textbf{\textcolor{violet}{0.765}} & \textbf{\textcolor{violet}{0.883}} & \textbf{\textcolor{violet}{0.983}} & \textbf{\textcolor{violet}{1.340}} & \textbf{\textcolor{violet}{1.975}}\\
\hline
\multirow{4}{*}{$\boldsymbol{\tau = 2}$} 
& \textbf{CRN} & 0.671	& 0.666 	& 0.741 	& 1.668 & 1.151 & 1.456 & 3.117 \\
& \textbf{CRN + SGA + RTM} & \textbf{\textcolor{blue}{0.629}} & \textbf{\textcolor{blue}{0.648}} & \textbf{\textcolor{blue}{0.674}} & \textbf{\textcolor{blue}{0.822}} &  \textbf{\textcolor{blue}{0.934}} & \textbf{\textcolor{blue}{0.948}} & \textbf{\textcolor{blue}{1.308}}\\
\cdashline{2-9}
& \textbf{CT} & 0.681 & 0.677	& 0.713 & 0.908 & 1.274 & 1.718 & 3.053 \\
& \textbf{CT + SGA + RTM} & \textbf{\textcolor{violet}{0.610}} & \textbf{\textcolor{violet}{0.673}}  & \textbf{\textcolor{violet}{0.688}} & \textbf{\textcolor{violet}{0.751}} & \textbf{\textcolor{violet}{0.786}} & \textbf{\textcolor{violet}{0.911}} & \textbf{\textcolor{violet}{0.983}}\\
\hline
\multirow{4}{*}{$\boldsymbol{\tau = 3}$} 
& \textbf{CRN} & 0.700 & 0.692 & 0.818 & 1.959 & 1.360 & 1.684 & 3.451\\
& \textbf{CRN + SGA + RTM} & \textbf{\textcolor{blue}{0.643}} & \textbf{\textcolor{blue}{0.666}} & \textbf{\textcolor{blue}{0.706}} & \textbf{\textcolor{blue}{0.862}} & \textbf{\textcolor{blue}{0.981}} & \textbf{\textcolor{blue}{0.997}} & \textbf{\textcolor{blue}{1.404}}\\
\cdashline{2-9}
& \textbf{CT} & 0.703 & 0.712 & 0.770 & 1.010 & 1.536 & 1.948 & 3.367\\
& \textbf{CT + SGA + RTM} & \textbf{\textcolor{violet}{0.630}} & \textbf{\textcolor{violet}{0.691}} & \textbf{\textcolor{violet}{0.718}} &  \textbf{\textcolor{violet}{0.805}} & \textbf{\textcolor{violet}{0.858}} &  \textbf{\textcolor{violet}{1.023}} & \textbf{\textcolor{violet}{1.127}} \\
\hline
\multirow{4}{*}{$\boldsymbol{\tau = 4}$} 
& \textbf{CRN} & 0.734 & 0.722 & 0.898 & 2.201 & 1.573 & 1.802 & 3.594\\
& \textbf{CRN + SGA + RTM} & \textbf{\textcolor{blue}{0.662}} & \textbf{\textcolor{blue}{0.681}} & \textbf{\textcolor{blue}{0.740}} & \textbf{\textcolor{blue}{0.922}} & \textbf{\textcolor{blue}{1.016}} & \textbf{\textcolor{blue}{1.060}}  & \textbf{\textcolor{blue}{1.478}} \\
\cdashline{2-9}
& \textbf{CT} & 0.726 & 0.748 & 0.822 & 1.089 & 1.762 & 2.095 & 3.516 \\
& \textbf{CT + SGA + RTM} & \textbf{\textcolor{violet}{0.650}} & \textbf{\textcolor{violet}{0.727}} &  \textbf{\textcolor{violet}{0.748}} &  \textbf{\textcolor{violet}{0.848}} & \textbf{\textcolor{violet}{0.919}} & \textbf{\textcolor{violet}{1.107}} & \textbf{\textcolor{violet}{1.231}} \\
\hline
\multirow{4}{*}{$\boldsymbol{\tau = 5}$} 
& \textbf{CRN} & 0.769 &	0.755 &	0.976 & 2.361 & 1.730 & 1.870 & 3.560 \\
& \textbf{CRN + SGA + RTM} & \textbf{\textcolor{blue}{0.682}} & \textbf{\textcolor{blue}{0.700}} & \textbf{\textcolor{blue}{0.771}}  &  \textbf{\textcolor{blue}{0.983}} &  \textbf{\textcolor{blue}{1.046}} & \textbf{\textcolor{blue}{1.118}} & \textbf{\textcolor{blue}{1.497}}\\
\cdashline{2-9}
& \textbf{CT} & 0.756 & 0.786	& 0.870 & 1.154 & 1.922 & 2.168 & 3.570 \\
& \textbf{CT + SGA + RTM} & \textbf{\textcolor{violet}{0.677}} & \textbf{\textcolor{violet}{0.761}} & \textbf{\textcolor{violet}{0.779}} &  \textbf{\textcolor{violet}{0.889}} & \textbf{\textcolor{violet}{0.995}} & \textbf{\textcolor{violet}{1.162}} & \textbf{\textcolor{violet}{1.315}}\\
\hline
\multirow{4}{*}{$\boldsymbol{\tau = 6}$} 
& \textbf{CRN} & 0.807 & 0.790 & 1.047 & 2.480 & 1.827 & 1.859 & 3.372\\
& \textbf{CRN + SGA + RTM} & \textbf{\textcolor{blue}{0.701}}   & \textbf{\textcolor{blue}{0.715}} & \textbf{\textcolor{blue}{0.792}} & \textbf{\textcolor{blue}{1.031}} & \textbf{\textcolor{blue}{1.064}} & \textbf{\textcolor{blue}{1.170}} & \textbf{\textcolor{blue}{1.480}}\\
\cdashline{2-9}
& \textbf{CT} & 0.789 &  0.821 &  0.909 &  1.205 &  2.052 & 2.184 & 3.436\\
& \textbf{CT + SGA + RTM} & \textbf{\textcolor{violet}{0.705}} & \textbf{\textcolor{violet}{0.800}} & \textbf{\textcolor{violet}{0.801}} & \textbf{\textcolor{violet}{0.935}} & \textbf{\textcolor{violet}{1.058}} & \textbf{\textcolor{violet}{1.200}} & \textbf{\textcolor{violet}{1.376}}\\
\hline
\end{tabular}
}
\label{tab:fully_synthetic_single_sliding_treatment}
\end{table}

\subsection{Experiments on Semi-Synthetic Data}
\label{sec:experiments_semi_syn}
We further validate our framework on a semi-synthetic dataset based on real-world medical data from intensive care units. This dataset is generated following the approach of \citet{melnychuk2022causal}, which builds upon the MIMIC-III dataset \citep{johnson2016mimic} to simulate patient trajectories with outcomes that reflect both endogenous and exogenous dependencies while incorporating treatment effects. 

\textbf{Results. } Table~\ref{tab:semi_synthetic} shows our framework \textbf{consistently improves performance} over the vanilla CRN baseline. The gains are more modest, potentially because confounding might be less prominent in this specific semi-synthetic setup (also shown by \citet{melnychuk2022causal}). Details are in Appendix~\ref{sup:exp_semi_synthetic}.

\begin{table}[t]
\caption{RMSE for one-step-ahead and $\tau$-step-ahead predictions on semi-synthetic data based on real-world medical data (MIMIC-III).}
\centering
\vspace{1mm} 
\scalebox{0.6}{
\large
\begin{tabular}{|c|c|c|c|c|c|c|c|c|c|c|}
\hline
  & $\boldsymbol{\tau=1}$ & $\boldsymbol{\tau=2}$ & $\boldsymbol{\tau=3}$ & $\boldsymbol{\tau=4}$ & $\boldsymbol{\tau=5}$ & $\boldsymbol{\tau=6}$ & $\boldsymbol{\tau=7}$ & $\boldsymbol{\tau=8}$ & $\boldsymbol{\tau=9}$ & $\boldsymbol{\tau=10}$ \\ \hline
\textbf{MSMs} & 0.37 & 0.57 & 0.74 & 0.88  & 1.14 & 1.95 & 3.44  & $> 10.0$ & $> 10.0$  & $> 10.0$ \\ 
\textbf{RMSNs} & 0.24 & 0.47 & 0.60 & 0.70 & 0.78 & 0.84 & 0.89 & 0.94  & 0.97 & 1.00 \\ 
\textbf{G-Net} & 0.34 & 0.67 & 0.83 & 0.94  & 1.03 & 1.10 & 1.16 & 1.21 & 1.25 & 1.29 \\ 
\textbf{CRN} & 0.30  & 0.48 & 0.59 & 0.65 & 0.68 & 0.71 & 0.72 & 0.74 & 0.76  & 0.78 \\ 
\cdashline{1-11}
\textbf{CRN + SGA + RTM} & \textbf{0.27}& \textbf{0.43} & \textbf{0.52}  & \textbf{0.58} &\textbf{0.62} & \textbf{0.65} & \textbf{0.67} & \textbf{0.69} & \textbf{0.72} & \textbf{0.73} \\
\hline
\end{tabular}
}
\label{tab:semi_synthetic}
\end{table}

\subsection{Ablation Studies}
\label{sec:ablation_studies_summary}
We run ablation experiments to evaluate the individual contributions of \textbf{SGA} and \textbf{RTM}. 

\textbf{Effectiveness of SGA. } The left panel of Table~\ref{tab:ablation_table} demonstrates that incorporating SGA \textbf{alone} consistently improves performance over baseline models. The improvements are particularly evident in settings with \textbf{higher} confounding levels (larger $\gamma$). This supports our claim that SGA's fine-grained sub-treatment group alignment leads to more effective deconfounding.

\textbf{Effectiveness of RTM.}  The right panel of Table~\ref{tab:ablation_table} shows that adding RTM consistently improves model performance. RTM's impact is more pronounced on \textbf{longer-term} predictions (larger $\tau$) and under \textbf{stronger} confounding (larger $\gamma$), confirming that RTM promotes better temporal generalization.

\vspace{-3mm}

\begin{table}[h]
\caption{Normalized RMSE for $\tau$-step-ahead predictions on fully-synthetic data, CRN/CT vs. CRN/CT+SGA (left) and CRN/CT vs. CRN/CT+RTM (right).}
\vspace{1mm} 
\label{tab:ablation_table}
\centering
\begin{subtable}[t]{0.48\textwidth}
\centering
\resizebox{\textwidth}{!}{
\begin{tabular}{|c|c|c|c|c|c|c|}
\hline
$\boldsymbol{\gamma}$ & \textbf{Method}  & $\boldsymbol{\tau = 3}$ & $\boldsymbol{\tau = 4}$ & $\boldsymbol{\tau = 5}$ & $\boldsymbol{\tau = 6}$ \\
\hline
\multirow{4}{*}{$\boldsymbol{ 2}$} 
& \textbf{CRN}  & 0.818  & 0.898 & 0.976 &	1.047 \\
& \textbf{CRN + SGA}  &  \textbf{\textcolor{blue}{0.698}} & \textbf{\textcolor{blue}{0.743}} & \textbf{\textcolor{blue}{0.782}}  & \textbf{\textcolor{blue}{0.810}} \\
\cdashline{2-7} 
& \textbf{CT}  & 0.770 & 0.822 & 0.870  & 0.909	\\
& \textbf{CT + SGA}  & \textbf{\textcolor{violet}{0.762}} & \textbf{\textcolor{violet}{0.813}}  & \textbf{\textcolor{violet}{0.854}}  & \textbf{\textcolor{violet}{0.876}} \\
\hline
\multirow{4}{*}{$\boldsymbol{ 6}$} 
& \textbf{CRN} & 3.451  & 3.594 & 3.560 & 3.372 	\\
& \textbf{CRN + SGA} &  \textbf{\textcolor{blue}{3.157}} & \textbf{\textcolor{blue}{3.284}} & \textbf{\textcolor{blue}{3.231}}  & \textbf{\textcolor{blue}{3.034}} \\
\cdashline{2-7}
& \textbf{CT}  & 3.367 & 3.516 & 3.570 & 3.436	\\
& \textbf{CT + SGA} & \textbf{\textcolor{violet}{3.231}} & \textbf{\textcolor{violet}{3.345}}  & \textbf{\textcolor{violet}{3.238}}  & \textbf{\textcolor{violet}{3.014}} \\
\hline
\end{tabular}
}
\end{subtable}
\hfill
\begin{subtable}[t]{0.48\textwidth}
\centering
\resizebox{\textwidth}{!}{
\begin{tabular}{|c|c|c|c|c|c|c|}
\hline
$\boldsymbol{\gamma}$ & \textbf{Method}  & $\boldsymbol{\tau = 3}$ & $\boldsymbol{\tau = 4}$ & $\boldsymbol{\tau = 5}$ & $\boldsymbol{\tau = 6}$ \\
\hline
\multirow{4}{*}{$\boldsymbol{2}$} 
& \textbf{CRN}  & 0.818  & 0.898 & 0.976 &	1.047 \\
& \textbf{CRN + RTM}  &  \textbf{\textcolor{blue}{0.791}} & \textbf{\textcolor{blue}{0.862}} & \textbf{\textcolor{blue}{0.907}}  & \textbf{\textcolor{blue}{0.934}} \\
\cdashline{2-7} 
& \textbf{CT}  & 0.770 & 0.822 & 0.870  & 0.909	\\
& \textbf{CT + RTM}  & \textbf{\textcolor{violet}{0.720}} & \textbf{\textcolor{violet}{0.752}}  & \textbf{\textcolor{violet}{0.787}}  & \textbf{\textcolor{violet}{0.819}} \\
\hline
\multirow{4}{*}{$\boldsymbol{ 6}$} 
& \textbf{CRN} & 3.451  & 3.594 & 3.560 & 3.372 	\\
& \textbf{CRN + RTM} &  \textbf{\textcolor{blue}{2.316}} & \textbf{\textcolor{blue}{2.469}} & \textbf{\textcolor{blue}{2.541}}  & \textbf{\textcolor{blue}{2.523}} \\
\cdashline{2-7}
& \textbf{CT}  & 3.367 & 3.516 & 3.570 & 3.436	\\
& \textbf{CT + RTM} & \textbf{\textcolor{violet}{1.883}} & \textbf{\textcolor{violet}{2.009}}  & \textbf{\textcolor{violet}{2.039}}  & \textbf{\textcolor{violet}{2.005}} \\
\hline
\end{tabular}
}
\end{subtable}
\end{table}

\vspace{-2mm}


\subsection{Analysis of Masking Strategies}
\label{sec:masking_strategies_summary}

To evaluate our choice of Gaussian noise for RTM, we compare it against alternative masking strategies: ($i$) \emph{masking with zeros} and ($ii$) \emph{interpolation-masking} that replaces input covariates from selected time-point $t$ with the {\bf average} covariates of $t-1$ and $t+1$. We observe the followings: 

\textbf{O1. Gaussian noise significantly outperforms other choices.} We hypothesize that this is because Gaussian noise acts as a more effective regularizer. It discourages the model from overfitting to specific input values at the masked time steps and forces it to learn more generalizable representations 
\begin{table}[ht]
\centering
\caption{Comparison of masking strategies on fully-synthetic data ($\gamma$=6) with normalized RMSE. Gaussian noise refers to CT+RTM. Other strategies are also applied to the CT model architecture.}
\label{tab:masking_comparison_ct_concise_rebuttal}
\resizebox{.5\textwidth}{!}{
\begin{tabular}{|c|c|c|c|c|}
\hline
\textbf{Masking Strategy}   & $\boldsymbol{\tau=3}$ & $\boldsymbol{\tau=4}$ & $\boldsymbol{\tau=5}$ & $\boldsymbol{\tau=6}$ \\ \hline
Zero masking & 2.821 & 3.072 & 3.122 & 2.986   \\ 
Interpolation  & 3.435 & 3.556 & 3.602  & 3.480    \\
No Masking  & 3.367  & 3.516 & 3.570   & 3.436   \\ 
\hline
Gaussian noise  & \textbf{1.883}  & \textbf{2.009}  & \textbf{2.039}  & \textbf{2.005} \\ 
\hline
\end{tabular}
}
\end{table}
from the historical context.

\textbf{O2. Zero masking performs sub-optimally.} It introduces a strong, and often unnatural signal at zero. Models risk using these artificial zeros or misinterpreting them, which negatively affects the learning of true underlying patterns.

\textbf{O3. Interpolation yields the highest error.} It creates artificially smooth data. This obscures natural data variability and can potentially impose a deterministic bias, preventing the model from capturing true temporal dynamics.



\subsection{RTM Attention Weight Analysis}
\label{sec:rtm_attention_summary}

To investigate how RTM promotes learning of long-range dependencies, we show the softmaxed self-attention weights from the final attention layer of Causal Transformer model. Figure~\ref{fig:rtm_attention_heatmap} compares attention patterns for CT+RTM versus CT (no RTM) when predicting the outcome at the last time-point ($t=58$) under high confounding ($\gamma=6$). This analysis confirms RTM's success in shifting the model's reliance from current time step towards historical context, thereby promoting temporal generalization and the learning of long-term causal patterns, as Figure~\ref{fig:rtm_attention_heatmap} shows:
\begin{figure}[t]
    \centering
    \includegraphics[width=1.0\textwidth]{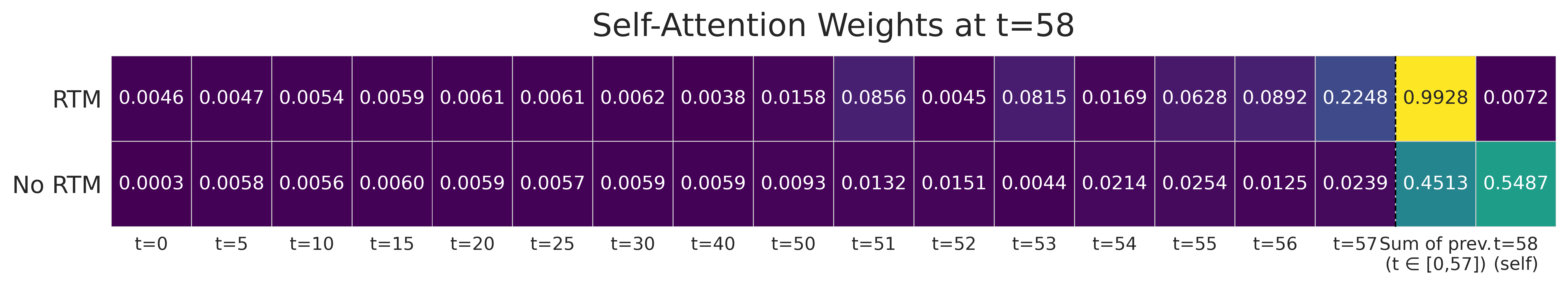}
    \caption{\textbf{Heatmap of self-attention weights.} \textbf{Left:} attention to past time points. \textbf{Rightmost columns:} sum of total past attention (\texttt{Sum of prev.}), and attention to the current time point (\texttt{t=58}).}
    \label{fig:rtm_attention_heatmap}
\end{figure}
\begin{itemize}[left=0pt,itemsep=0em, topsep=0em]
    \item \textbf{With RTM, the model allocates over 99\% of its attention to previous time-points}. This distributed attention across historical data demonstrates effective leveraging of past information.
    \item \textbf{Without RTM, attention to past steps is significantly lower,} with the model focusing over 54\% of its attention on the current time-point. This highlights the tendency of models without RTM to \textbf{over-rely on information at the current time step} rather than learn long-term causal patterns.
\end{itemize}

\subsection{Clustering Sensitivity}
\label{sec:clustering_sensitivity_summary}

In our experiments, SGA uses Gaussian Mixture Models (GMMs) for \emph{treatment‑agnostic} clustering to identify sub-treatment groups. We analyze SGA's sensitivity to the choice of clustering algorithm and the number of clusters (K). Specifically, we compare GMM with k-means using varying K.

Evaluations are performed on the fully-synthetic dataset under the challenging setting of \textbf{long-term predictions} ($\tau$=6) and \textbf{higher confounding} ($\gamma$=6). We observe from Table~\ref{tab:clusters_sensitivity} : 
    
\textbf{O1.} SGA's performance \textbf{remains relatively stable} across different numbers of clusters, suggesting low sensitivity to this hyperparameter.

\textbf{O2.} \textbf{Both k-means and GMM yield comparable performance}, indicating robustness to the specific clustering algorithm used.

\vspace{-4mm}

\begin{table}[h]
\centering
\caption{Sensitivity analysis of SGA to the choice of clustering algorithm and the number of clusters (K). The table reports the normalized RMSE with $\gamma=6$ at $\tau=6$.}
\vspace{1mm} 
\label{tab:clusters_sensitivity}
\scalebox{0.8}{
\begin{tabular}{|c|c|c|cccc|}
\hline
\multirow{2}{*}{\textbf{Model}} & \multirow{2}{*}{\textbf{Clustering Algorithm}} 
& \textbf{No SGA} & \multicolumn{4}{c|}{\textbf{SGA with varying $K$}} \\
\cline{4-7}
& & (K=1) & K=2 & K=3 & K=5 & K=10 \\
\hline
\multirow{2}{*}{\textbf{CRN+SGA}} & \textbf{k-means} & 3.372 & 3.568 & 3.301 & \textbf{\textcolor{black}{3.259}} & 3.504\\
 & \textbf{GMM}  & 3.372 & \textbf{\textcolor{black}{3.034}}  &  3.258 & 3.276 & 3.595 \\ \hline
\multirow{2}{*}{\textbf{CT+SGA}}  & \textbf{k-means} &  3.436 & 3.154  &  \textbf{\textcolor{black}{3.008}} & 3.206 & 3.059\\
 & \textbf{GMM}  & 3.436 & \textbf{\textcolor{black}{3.014}} &  3.423 & 3.158 & 3.082 \\ \hline
\end{tabular}
}
\end{table}

These findings confirm SGA's {\bf robustness} to the clustering algorithm and the number of clusters, making it practical for diverse real-world settings where the optimal number of clusters may be unknown or require tuning.



\section{Conclusion}

We address the critical challenge of counterfactual outcome estimation in time series by introducing two novel, synergistic approaches: Sub-treatment Group Alignment (SGA) and Random Temporal Masking (RTM). SGA tackles time-varying confounding at each time point by aligning fine-grained sub-treatment group distributions, leading to tighter counterfactual error bound and more effective deconfounding. Complementarily, RTM promotes temporal generalization and robust learning from historical patterns by randomly masking covariates, encouraging the model to preserve underlying causal relationships across time steps and rely less on potentially noisy contemporaneous time steps.


Our comprehensive experiments demonstrate that SGA and RTM are broadly applicable and significantly enhance existing state-of-the-art methods. While each approach individually improves performance, their synergistic combination consistently achieves SOTA performance on both synthetic and semi-synthetic benchmark datasets. Together, SGA and RTM offer a flexible and effective framework for improving causal inference from observational time series data.


\bibliography{neurips_2025}
\bibliographystyle{abbrvnat}

\newpage

\appendix

\section{Potential Outcomes Framework with Time-Varying Treatments and Outcomes}

\label{sec:potential_outcome}
Building on the potential outcomes framework \citep{rosenbaum1983central, rubin2005causal}, we extend these assumptions to accommodate time-varying treatments and outcomes, following \citet{robins2008estimation}. 

\begin{assumption}
\label{identify_a1}
(Consistency) If $\mathbf{\bar{A}}_t = \mathbf{\bar{a}}_t$ is a given sequence of treatments for some patient, then $\mathbf{Y}_{t+1} [\mathbf{\bar{a}}_t] = \mathbf{Y}_{t+1}$
This means an individual's potential outcome under the observed treatment history is the observed outcome.
\end{assumption}

\begin{assumption}
\label{identify_a2}
(Sequential Positivity) Positivity states that there is non-zero probability or not receiving any of the counterfactual treatment. It can be expressed as $0 < P(\mathbf{A}_t =\mathbf{a}_t | \bar{\mathbf{H}}_t  = \bar{\mathbf{h}}_t) < 1$, if $P(\bar{\mathbf{H}}_t  = \bar{\mathbf{h}}_t)>0$.
\end{assumption}

\begin{assumption}
\label{identify_a3}
(Sequantial Ignorability) There is no unobserved confounding of treatment at any time and any future outcome. This can be expressed as \textcolor{black}{$\mathbf{A}_t \independent  \mathbf{Y}_{t+1} [\mathbf{a}_t] |  \bar{\mathbf{H}}_t, \forall \; \mathbf{a}_t \in \mathcal{A}$.}
\end{assumption}

Using assumptions \ref{identify_a1}--\ref{identify_a3}, \citet{robins1986new} establishes the sufficient conditions for identifiability through G-computation, ensuring that causal effects can be appropriately identified.

Fig~\ref{causal_graph} visualizes Causal Directed Acyclic Graphs (DAGs) illustrating causal relationships. 

\begin{figure}[h]
\centering
\begin{subfigure}[t]{0.47\textwidth}
    \centering
    \begin{tikzpicture}
    \node[state,circle,fill=black!15] (1) {$A$};
    \node[circle,draw,inner sep=4] (2) [above right =of 1, yshift=-0.5cm, xshift=0.3cm] {$X$};
    \node[state,circle,fill=black!15] (3) [right =of 1, xshift=1.2cm, xshift=0.6cm] {$Y$};
    \path (2) edge (1);
    \path (2) edge (3);
    \path (1) edge (3);
    \end{tikzpicture}
    \caption*{(a) Static (non-time-series) Scenario}
\end{subfigure}
\\
\textit{In the static (non-time-series) scenario, we have $A$ as the treatment assignment, $X$ as the covariate, and $Y$ as the outcome.}
\begin{subfigure}[t]{1.0\textwidth}
    \centering
    \begin{tikzpicture}
    \node[state,circle,minimum size=1.2cm,scale=0.6] (1) {$X_{t-1} \cup V$};
    \node[state,circle,minimum size=1.2cm,scale=0.7] (2) [right =of 1, xshift=3cm] {$X_{t} \cup  V $};
    \node[state,circle,fill=black!15, minimum size=1.2cm,scale=0.8] (3) [below =of 1] {$A_{t-1}$};
    \node[state,circle,fill=black!15, minimum size=1.2cm,scale=0.8] (4) [below =of 2] {$A_{t}$};
    \node[circle,draw,fill=black!15,minimum size=1.2cm,scale=0.8] (5) [below =of 3, xshift=-3cm] {$Y_{t-1}$};
    \node[circle,draw,fill=black!15,minimum size=1.2cm,scale=0.8] (6) [below =of 3, xshift=3.0cm] {$Y_{t}$};
    \node[circle,draw,fill=black!15,minimum size=1.2cm,scale=0.8] (7) [below =of 4, xshift=3.0cm] {$Y_{t+1}$};
    \path (1) edge (2);
    \path (1) edge (3);
    \path (1) edge (4);
    \path (1) edge (6);
    \path (1) edge[bend right=10]  (7);
    \path (2) edge (4);
    \path (2) edge[bend left=20] (7);
    \path (3) edge (2);
    \path (3) edge (6);
    \path (3) edge (7);
    \path (3) edge (4);
    \path (4) edge (7);
    \path (5) edge (6);
    \path (5) edge[bend right=20] (7);
    \path (5) edge[bend left=20] (1);
    \path (5) edge[out=100, in=120, distance=3.2cm] (2);
    \path (5) edge (3);
    \path (5) edge (4);
    \path (6) edge (2);
    \path (6) edge (4);
    \path (6) edge (7);
    \end{tikzpicture}
    \caption*{(b) Time-Series Scenario}
\end{subfigure}
\\
\textit{In the time-series scenario, $A_t$ is the treatment at time t, $X_t \cup V$ represents observed covariates at time $t$, and $Y_t$ is the outcome at time $t$. Here, $V$ denotes static covariates that do not change over time. The diagrams capture the dynamics of treatment effects over time, showing how each component influences subsequent outcomes within the causal framework.}
\caption{\textbf{Causal Directed Acyclic Graphs (DAGs) Illustrating Causal Relationships.} \textbf{(a)} demonstrate a static (non-time-series) scenario. \textbf{(b)} illustrates a time-series scenario.}
\label{causal_graph}
\end{figure}

\section{Related Work}
\label{sec:complete_related_work}

\textbf{Estimating counterfactual outcomes under static scenarios.} Many methods have been proposed to learn a \emph{balanced} representation that aligns the distributions across treatment groups, effectively addressing confounding in static settings. A foundational work in this area, CFRNet proposed by \citet{shalit2017estimating}, establishes a counterfactual error bound illustrating that the expected error in estimating individual treatment effects (ITE) is bounded by the sum of its standard generalization error and the discrepancy between treatment group distributions induced by the representation. This concept has been further explored in several subsequent studies on deep causal inference \citep{yao2018representation, kallus2020deepmatch, du2021adversarial}. However, these methods primarily focus on static data, and their approach of aligning overall treated and control group distributions may not sufficiently adaptable to time-series data~\citep{hernan2000marginal,mansournia2012effect}, where time-dependent confounders make it difficult to disentangle the true effect of a treatment from these caused by the confounding variables.


\textbf{Estimating counterfactual outcomes over time.} Estimating counterfactual outcomes in time-series data is challenging due to time-varying confounders. Traditional methods such as G-computation and marginal structural models \citep{robins1986new, robins2000marginal, hernan2001marginal, robins2008estimation,xu2016bayesian} often lack flexibility for complex datasets and rely on strong assumptions. To address these limitations, researchers have developed models that build on the potential outcomes framework initially proposed by Rubin \citep{rubin1978bayesian} and extended to time series by \citet{robins2008estimation}. Notable among recent methods are Recurrent Marginal Structural Networks (RMSNs) \citep{lim2018forecasting}, G-Net \citep{li2020g}, Counterfactual Recurrent Networks (CRN) \citep{bica2020estimating}, and the Causal Transformer (CT) \citep{melnychuk2022causal}, which use approaches such as propensity networks and adversarial learning to mitigate the effects of time-varying confounding. The CRN employs recurrent neural networks like LSTMs, while the CT uses Transformer-based architectures, representing the state-of-the-art in this domain. However, practical challenges with adversarial training can affect the stability of causal effect estimations. Specifically, training adversarial networks can be challenging due to issues such as mode collapse and oscillations \citep{liang2018generative}. Additionally, adversarial training minimizes the Jensen-Shannon divergence (JSD) only when the discriminator is optimal \citep{arjovsky2017towards}, which may not always be achievable in practice; even when the discrminator is optimal, using JSD optimizing relatively loose upper bounds on the counterfactual error. To address these challenges, we propose using the Wasserstein-1 distance. The Wasserstein distance is bounded above by the Kullback-Leibler divergence (JSD is a symmetrized and smoothed version of the Kullback–Leibler divergence) and provides stronger theoretical guarantees \citep{redko2017theoretical,mansour2012multiple}. Moreover, the Wasserstein distance has stable gradients even when the compared distributions are far apart \citep{gulrajani2017improved}, which enhances training stability and effectiveness.

\textbf{Masked language modeling.} Masked language modeling (MLM) is a common self-supervised pre-training technique for large language models. It operates by randomly masking certain words or tokens in the input, with the model trained to predict the masked tokens. BERT \citep{devlin2018bert} is the most well-known model that employs this technique. Recent studies have also demonstrated the effectiveness of MLM in enhancing generalization across various sequence-based tasks. For example, \citet{chaudhary2020dict} showed that when combined with cross-lingual dictionaries, MLM not only improves predictions for the original masked word but also generalizes to its cross-lingual synonyms. Additionally, \citet{czinczoll2024nextlevelbert} illustrated how MLM enhances generalization in long-document tasks by leveraging higher-level semantic representations. Inspired by the success of masking strategies in language models, we introduce Random Temporal Masking (RTM) for time-series data. Unlike MLM, which focuses on predicting the masked inputs, RTM encourages the model to focus on information that is crucial not only for the current time point but also for future time points, preserve causal information, and reduce the risk of overfitting to factual outcomes.

\section{Theorems and Proofs}
\label{sup_def}

\begin{definition}[Definition A4 in \citet{shalit2017estimating}]
Let $\Phi : \mathcal{X} \to \mathcal{R}$ be a representation function. Let $h : \mathcal{R} \times \{0, 1\} \to \mathcal{Y}$ be an hypothesis defined over the representation space $\mathcal{R}$. The expected loss for the unit and treatment pair $(x, t)$ is:
\[
\ell_{h, \Phi}(x, t) = \int_{\mathcal{Y}} L(Y_t, h(\Phi(x), t)) p(Y_t | x) dY_t,
\]
where $L(\cdot,\cdot)$ is a loss function, from $\mathcal{Y} \times \mathcal{Y}$ to $\mathbb{R}_{+}$.
\end{definition}

\begin{definition}[Definition A5 in \citet{shalit2017estimating}]
\label{def:f_cf}
The expected factual loss and counterfactual losses of $h$ and $\Phi$ are, respectively:
\[
\epsilon_F(h, \Phi) = \int_{\mathcal{X} \times \{0, 1\}} \ell_{h, \Phi}(x, t) p(x, t) dx dt,
\]
\[
\epsilon_{CF}(h, \Phi) = \int_{\mathcal{X} \times \{0, 1\}} \ell_{h, \Phi}(x, t) p(x, 1 - t) dx dt,
\] 
where $p(x, t)$ is distribution over $\mathcal{X} \times \{0, 1\}$ \\
 \end{definition}

\begin{definition}
\label{def:lipschitz}
    For some $K \geq 0$, the set of $K$-Lipschitz functions denotes the set of functions $f$ that verify: 
    $$
    \|f(x) - f(x')\| \leq K \|x-x'\|, \; \forall x,x' \in \mathcal{X}
    $$
 \end{definition}
Here, we assume that the hypothesis class $\mathbb{H}$ is a subset of $\lambda_{H}$-Lipschitz functions, where $\lambda_{H}$ is a positive constant, and we assume that the true labeling functions are $\lambda$-Lipschitz for some positive real number $\lambda$. Also if $f$ is differentiable, then a sufficient condition for $K$-Lipschitz constant is if $\|\frac{\partial f}{\partial x}\| \leq x$ for all $s \in \mathcal{X}$. \\

\begin{assumption}[Assumption A2 in \citet{shalit2017estimating}]
There exists a constant $K > 0$ such that for all $x \in \mathcal{X}, t \in \{0,1\}, \|\frac{\partial p(Y_t|x)}{\partial x}\| \leq K$. \\
\end{assumption}

\begin{assumption} [Assumption A3 in \citet{shalit2017estimating}]
\label{sup_assumption_A3}
The loss function $L$ is differentiable, and there exists a constant $K_L > 0$ such that $\left| \frac{dL(y_1,y_2)}{dy_i} \right| \leq K_L$ for $i = 1, 2$. Additionally, there exists a constant $M$ such that for all $y_2 \in \mathcal{Y}$, $M \geq \int_{\mathcal{Y}} L(y_1, y_2) dy_1$. \\
\end{assumption}

\begin{definition} [Definition A12 in \citet{shalit2017estimating}]
Let $\frac{\partial \Phi(x)}{\partial x}$ be the Jacobian matrix of $\Phi$ at point $x$, i.e., the matrix of the partial derivatives of $\Phi$. Let $\sigma_{\max}(A)$ and $\sigma_{\min}(A)$ denote respectively the largest and smallest singular values of a matrix $A$. Define $\rho(\Phi) = \frac{\sup_{x \in \mathcal{X}} \sigma_{\max}(\frac{\partial \Phi(x)}{\partial x})}{\sigma_{\min}(\frac{\partial \Phi(x)}{\partial x})}$. \\
\end{definition}

\begin{definition} [Definition A13 in \citet{shalit2017estimating}]
We will call a representation function $\Phi: \mathcal{X} \rightarrow \mathcal{R}$ Jacobian-normalized if $\sup_{x \in \mathcal{X}} \sigma_{\max}(\frac{\partial \Phi(x)}{\partial x}) = 1$ 
\end{definition}
\begin{remark}
    Any non-constant representation function $\Phi$ can be Jacobian-normalized by a simple scalar multiplication.
\end{remark}

\begin{lemma} [Lemma A3 in \citet{shalit2017estimating}]
Let $u = p(t=1)$, then we have,
\[
\epsilon_F(h, \Phi) = u \cdot \epsilon_{F}^{t=1}(h, \Phi) + (1 - u) \cdot \epsilon_{F}^{t=0}(h, \Phi)
\]
\[
\epsilon_{CF}(h, \Phi) = (1 - u) \cdot \epsilon_{CF}^{t=1}(h, \Phi) + u \cdot \epsilon_{CF}^{t=0}(h, \Phi)
\] \\
\end{lemma}

\begin{definition}
\label{def:new_error_f}
 Let $u = p(t=1)$ be the marginal probability of treatment and assume $0 < u < 1$.
 \begin{equation*}
     \epsilon_F^{\star}(h, \Phi) = (1 - u) \epsilon_{F}^{t=1}(h, \Phi) + u \epsilon_{F}^{t=0}(h, \Phi)  \\
 \end{equation*}
\end{definition}

\begin{lemma}[Decomposition over sub‑groups]
\label{lem:subgroup_decomp}
Assume the representation $\Phi$ induces $K$ treatment–agnostic clusters
so that each mixture weight is shared, i.e., $w_k^0=w_k^1=w_k$ with
$\sum_{k=1}^{K}w_k=1$.  
Define the sub‑domain factual and counterfactual losses
\[
\epsilon_F^{k}(h,\Phi)\;=\;\int_{\mathcal X\times\{0,1\}}\!
        \ell_{h,\Phi}(x,t)\,p_k(x,t)\,dx\,dt, 
\]
\[
\epsilon_{CF}^{k}(h,\Phi)\;=\;\int_{\mathcal X\times\{0,1\}}\!
        \ell_{h,\Phi}(x,t)\,p_k(x,1-t)\,dx\,dt,
\]
where $p_k(x,t)$ is the conditional distribution of $(X,T)$ for sub-group
$k$.  
Then
\[
\epsilon_F(h,\Phi)\;=\;\sum_{k=1}^{K} w_k\,\epsilon_F^{k}(h,\Phi),
\qquad
\epsilon_{CF}(h,\Phi)\;=\;\sum_{k=1}^{K} w_k\,\epsilon_{CF}^{k}(h,\Phi).
\]
\end{lemma}

Now using the Definition~\ref{def:new_error_f}, we rewrite Lemma A8 from~\citet{shalit2017estimating}. \\

\begin{theorem}[Lemma A8 from~\citet{shalit2017estimating}]
\label{thm:complete_theorem_1}
    Let $u = p(t=1)$ be the marginal probability of treatment and assume $0 < u < 1$. Let $\Phi:\mathcal{X} \rightarrow \mathcal{R}$ be a one-to-one and Jacobian-normalized representation function. Let $K$ be the Lipschitz constant of the functions $p(Y_t|x)$ on $\mathcal{X}$. Let $K_L$ be the Lipschitz constant of the loss function $L$, and M be as in Assumption \ref{sup_assumption_A3}. Let $h : R \times \{0, 1\} \rightarrow Y$ be an hypothesis with Lipschitz constant bK:
\begin{equation}
\epsilon_{CF}(h, \Phi) \leq \epsilon_F^{\star}(h, \Phi) +
2 \left(M \rho(\Phi) + b\right) \cdot K \cdot K_L \cdot W_1(p_{\Phi}^{0}, p_{\Phi}^{1}),
\end{equation}
where $B_\Phi = \left(M \rho(\Phi) + b\right) \cdot K \cdot K_L$ is a constant and $p_{\Phi}^{a}$ is the distribution of the random variable $\Phi(X)$ conditioned on $A=a$, that is, representations for individuals receiving treatment $a \in \{0,1\}$.
\end{theorem}

\begin{theorem}[Simplified Lemma A8 from~\citet{shalit2017estimating}]
Let $\Phi:\mathcal{X} \rightarrow \mathcal{R}$ be a one-to-one and Jacobian-normalized representation function. Let $h : R \times \{0, 1\} \rightarrow Y$ be a hypothesis with Lipschitz constant:
\begin{equation}
\epsilon_{CF}(h, \Phi) \leq \epsilon_{F}^{\star}(h, \Phi) + 2 \cdot B_\Phi \cdot W_1(p_\Phi^{0}, p_\Phi^{1}),
\end{equation}
where $B_\Phi$ is a constant and $p_{\Phi}^{a}$ is the distribution of the random variable $\Phi(X)$ conditioned on $A=a$, that is, representations for individuals receiving treatment $a \in \{0,1\}$.
\end{theorem}

\begin{definition}
\label{def:wass}
\textbf{Wasserstein Distance.} The Kantorovich-Rubenstein dual representation of the Wasserstein-1 distance~\citep{villani2009optimal} between two distributions $p_{\Phi}^{0}$ and $p_{\Phi}^{1}$ is defined as
\begin{equation*}
\textstyle W_1(p_{\Phi}^{0},p_{\Phi}^{1}) = \sup_{\norm{f}_L \leq 1} \mathbb{E}_{x\sim p_{\Phi}^{0}}[f(x)] - \mathbb{E}_{x\sim p_{\Phi}^{1}}[f(x)],
\end{equation*}
where the supremum is over the set of $1$-Lipschitz functions (all Lipschitz functions $f$ with Lipschitz constant $L \le 1$. For notational simplicity, we use $D(X_1,X_2)$ to denote a distance between the distributions of any pair of random variables $X_1$ and $X_2$. For instance, $W_1(\Phi(X_0),\Phi(X_1))$ denotes the Wasserstein-1 distance between the distributions of the random variables $\Phi(X_0)$ and $\Phi(X_1)$ for
any transformation $\Phi$. 
\end{definition}

Next, motivated by the generalization bound~\citet{liu2025understanding} in the field of domain adaptation , we show that \textbf{sub-treatment group alignment improves the original alignment method in Theorem~\ref{thm:theorem1} by optimizing a tighter upper bound of the counterfactual error}. 

\textbf{Theorem~\ref{thm:weight_subdomain_distance_stronger} [Complete]} (SGA Improves Generalization Bounds) \textit{Under the following assumptions:}

\textit{\textbf{A1.} For all $k$, the sub-distributions $P_{\Phi,k}^{0}$ and $P_{\Phi,k}^{1}$ are Gaussian distributions with means $m_k^{0}$ and $m_k^{1}$, and covariances $\Sigma_k^{0}$ and $\Sigma_k^{1}$, respectively. The distance between corresponding sub-distributions is less than or equal to the distance between non-corresponding sub-distributions, i.e., $W_1(P_{\Phi,k}^{0}, P_{\Phi,k}^{1}) \leq W_1(P_{\Phi,k}^{0}, P_{\Phi,k^{\prime}}^{1})$ for $k \neq k^{\prime}$.}

\textit{\textbf{A2.} The sub-treatment group weights are identical across treatment groups, i.e., $w_k^1 = w_k^0$ for all $k \in [K]$. Moreover, there exists a small constant $\epsilon >0$, such that ${\max_{k \in [K]}}(\text{tr} (\Sigma_k^{0})) \leq \epsilon$ and ${\max_{k\in[K]}}(\text{tr} (\Sigma_k^{1})) \leq \epsilon$.} 

\textit{The following inequalities hold:
\begin{equation*}
\begin{split}
&\epsilon_{CF}(h, \Phi) \le \epsilon_{F}^{\star}(h, \Phi)  + 2 B_\Phi (\textstyle \sum_{k=1}^K w_k^{1} W_1(P_{\Phi,k}^{0}, P_{\Phi,k}^{1})) \\
& \textstyle\sum_{k=1}^K w_k^{1} W_1(P_{\Phi,k}^{0}, P_{\Phi,k}^{1}) \leq W_1(p_\Phi^{0}, p_\Phi^{1}) + \delta_c,
\end{split}
\end{equation*}
where $B_{\Phi}$ is the same constant in Theorem~\ref{thm:theorem1} and $\delta_c$ is $4\sqrt{\epsilon}$.} 
\begin{proof}[Proof of Theorem~\ref{thm:weight_subdomain_distance_stronger}]
Corollary~\ref{cor:sub_A8} proves the first statement while Theorem~\ref{thm:proof_weight_subdomain_distance_stronger_inequation2} proves the second.
\end{proof}


\begin{definition}(Wasserstein-like distance between Gaussian Mixture Models)
\label{defwassm} 
Assume that both $X_0$ and $X_1$ are mixtures of $K$ sub-domains. In other words, we have $p_{\Phi}^{0} = \sum_{k=1}^{K} w_k^{0} P_{\Phi,k}^{0}$ and $p_{\Phi}^{1} = \sum_{k=1}^{K} w_k^{1} P_{\Phi,k}^{1}$ where for $a \in {0,1}$, $w_k^{a}$ represents the proportion of the $k$-th sub-distribution in treatment group $a$. $P_{\Phi,k}^{a}$ denotes the distribution of the representations in the $k$-th sub-group under treatment $a$. We define:
\begin{equation}
    MW_1(p_{\Phi}^{0},p_{\Phi}^{1}) = \min_{w \in \Pi(\mathbf{w^{0}},\mathbf{w^{1}})} \sum_{k=1}^K \sum_{k^{\prime}=1}^K w_{k,k^{\prime}} W_1(P_{\Phi,k}^{0},P_{\Phi,k^{\prime}}^{1})
\end{equation}
where $\mathbf{w^{0}}\doteq[w_1^{0},\dots,w_K^{0}]$ and $\mathbf{w^{1}}\doteq[w_1^{1},\dots,w_K^{1}]$ belong to $\Delta^{K}$ (the $K-1$ probability simplex). $\Pi(w^{0},w^{1})$ represents the simplex $\Delta^{K \times K}$ with marginals $\mathbf{w^{0}}$ and $\mathbf{w^{1}}$. \\
\end{definition}

\begin{lemma}[Extension to Lemma 4.1 of \citet{delon2020wasserstein}]
\label{w1_proof_lemma}
Let $\mu_0 = \sum_{k=1}^{K_0} \pi_0^k \mu_0^k$ with $\mu_0^k = \mathcal{N} (m_0^k, \Sigma_0^k)$ and $\mu_1 = \sum_{k=1}^{K_1} \pi_1^k \delta_{m_1^k}$. Let $\Tilde{\mu_0} =  \sum_{k=1}^{K_0} \pi_0^k \delta_{m_0^k}$ ($\Tilde{\mu_0}$ only retains the means of $\mu_0$). Then, \\
$$MW_1(\mu_0,\mu_1)  \leq W_1 (\Tilde{\mu_0},\mu_1) + \sum_{k=1}^{K_0} \pi_0^k \sqrt{\text{tr }(\Sigma_0^k)} $$
where $\mathbf{\pi_0}\doteq[\pi_0^1,\dots,\pi_0^k]$ and $\mathbf{\pi_1}\doteq[\pi_1^1,\dots,\pi_1^k]$ belong to $\Delta^{K}$ (the $K-1$ probability simplex)
\end{lemma}
\begin{proof}
\begin{equation}
\begin{split}
MW_1(\mu_0,\mu_1) & =\inf_{w \in \Pi(\pi_0,\pi_1)} \sum_{k,l} w_{k,l} W_1(\mu_0^k,\delta_{m_1^l}) \\
& \leq \inf_{w \in \Pi(\mathbf{\pi_0},\mathbf{\pi_1})} \sum_{k,l} w_{k,l} W_2(\mu_0^k,\delta_{m_1^l}) \\
& = \inf_{w \in \Pi(\mathbf{\pi_0},\mathbf{\pi_1})} \sum_{k,l} w_{k,l} \left[\sqrt{||m_1^l-m_0^k||^2 + \text{tr }(\Sigma_0^k)} \right] \\
& \leq \inf_{w \in \Pi(\mathbf{\pi_0},\mathbf{\pi_1})} \sum_{k,l} w_{k,l} ||m_1^l-m_0^k|| + \sum_{k} \pi_0^k \sqrt{\text{tr }(\Sigma_0^k)} \\
& \leq W_1 (\Tilde{\mu_0},\mu_1) + \sum_{k=1}^{K_0} \pi_0^k \sqrt{\text{tr }(\Sigma_0^k)} \\
\end{split}
\end{equation}
\end{proof}
\begin{remark}
We use $\mu_0$, $\mu_1$, and $\tilde{\mu_0}$ to represent a general scenario for measuring the distance between a Gaussian mixture and a mixture of Diract distributions. In the following proofs, we will utilize the defined notation. For instance, $\mu_0$ can be denoted as $p_{\Phi}^{0}$, while $\tilde{\mu_0}$ corresponds to $\tilde{p_{\Phi}^{0}}$.
\end{remark}

For completeness, we recall the following result, Theorem A.10 and Theorem A.11 of \citet{liu2025understanding}, which we rely on in our analysis.

\begin{theorem}[Theorem A.10 in \citep{liu2025understanding}]
\label{w1_proof}
Let $p_{\Phi}^{0}$ and $p_{\Phi}^{1}$ be two Gaussian mixtures with $p_{\Phi}^{0} = \sum_{k=1}^{K} w_k^{0} P_{\Phi,k}^{0}$ and $p_{\Phi}^{1} = \sum_{k=1}^{K} w_k^{1} P_{\Phi,k}^{1}$. For all $k$, $P_{\Phi,k}^{0}$ / $P_{\Phi,k}^{1}$ are Gaussian distributions with mean $m_k^0$ / $m_k^1$ and covariance $\Sigma_k^0$ / $\Sigma_k^1$. If for $\forall$ k, $k^{\prime}$, we assume there exists a small constant $\epsilon >0$, such that $\max_{k}(\text{trace} (\Sigma_k^0)) \leq \epsilon$ and $\max_{k^{\prime}}(\text{trace} (\Sigma_{k^\prime}^1)) \leq \epsilon$.
then:
\begin{equation}
    MW_1(p_{\Phi}^{0},p_{\Phi}^{1})  \leq W_1 (p_\Phi^{0}, p_\Phi^{1}) + 4\sqrt{\epsilon}
\end{equation}
\end{theorem}

\begin{proof}
Here, we follow the same structure of the proof for Wassertein-2 in \citet{delon2020wasserstein}. 
Let $(P_\phi^{0})^n_n$ and $((P_\phi^{1})^n_n$ be two sequences of mixtures of Dirac masses respectively converging to $P_\phi^{0}$ and $P_\phi^{1}$ in $\mathcal{P}_1 (\mathbb{R}^d)$. Since $MW_1$ is a distance,
$$
\begin{aligned}
MW_1 (P_\phi^{0}, P_\phi^{1}) & \leq MW_1 ((P_\phi^{0})^n, (P_\phi^{1})^n) + MW_1 (P_\phi^{0}, (P_\phi^{0})^n) + MW_1 (P_\phi^{1}, (P_\phi^{1})^n) \\ & = W_1 ((P_\phi^{0})^n, (P_\phi^{1})^n) + MW_1 (P_\phi^{0}, (P_\phi^{0})^n) + MW_1 (P_\phi^{1}, (P_\phi^{1})^n)
\end{aligned}
$$
We can study the limits of these three terms when n $\rightarrow +\infty$

First, observe that $MW_1 (P_\phi^{0}, P_\phi^{1})  =  W_1 ((P_\phi^{0})^n, (P_\phi^{1})^n) \underset{n \rightarrow +\infty}{\rightarrow} W_1 (P_\phi^{0},P_\phi^{1})$ since $W_1$ is continuous
on $\mathcal{P}_1 (\mathbb{R}^d)$.

Second, based on Lemma \ref{w1_proof_lemma}, we have that
\begin{equation}
MW_1 (P_\phi^{0}, (P_\phi^{0})^n) \leq W_1 (\Tilde{P_\phi^{0}}, (P_\phi^{0})^n) + \sum_{k=1}^K w_k^0 \sqrt{\text{ tr} (\Sigma_k^0)} \underset{{n \rightarrow +\infty}}{\rightarrow}  W_1 (\Tilde{P_\phi^{0}}, P_\phi^{0}) + \sum_{k=1}^K w_k^{0} \sqrt{\text{ tr} (\Sigma_k^0)} \nonumber\end{equation}
We observe that $x\mapsto \sqrt{x}$ is a concave function, thus by Jensen's inequality, we have that
$$
\sum_{k=1}^K w_k^0 \sqrt{\text{ tr} (\Sigma_k^0)} \leq \sqrt{\sum_{k=1}^K w_k^0 \text{ tr} (\Sigma_k^0)}
$$
Also By Jensen's inequality, we have that, $$W_1 (\Tilde{P_\phi^{0}}, P_\phi^{0}) \leq W_2 (\Tilde{P_\phi^{0}},P_\phi^{0}).$$
And from Proposition 6 in \citep{delon2020wasserstein}, we have
$$
 W_2 (\Tilde{P_\phi^{0}},P_\phi^{0}) \leq \sqrt{\sum_{k=1}^K w_k^0 \text{ tr} (\Sigma_k^0)}
$$
Similarly for $MW_1 (P_\phi^{1}, (P_\phi^{1})^n)$ the same argument holds. 
Therefore we have,
$$
    \lim_{n\rightarrow \infty} MW_1 (P_\phi^{0}, (P_\phi^{0})^n) \leq 2 \sqrt{\sum_{k=1}^K w_k^0 \text{ tr} (\Sigma_k^0)} 
$$
And 
$$
    \lim_{n\rightarrow \infty} MW_1 (P_\phi^{1}, (P_\phi^{1})^n) \leq 2 \sqrt{\sum_{k=1}^K w_k^1 \text{ tr} (\Sigma_k^1)} 
$$
We can conclude that:
\begin{equation*}
    \begin{split}
        MW_1 (P_\phi^{0}, P_\phi^{1}) & \leq \lim \inf_{n \rightarrow \infty} (W_1 ((P_\phi^{0})^n, (P_\phi^{1})^n) + MW_1 (P_\phi^{0}, (P_\phi^{0})^n) + MW_1 (P_\phi^{1}, (P_\phi^{1})^n)) \\
        & \leq W_1 (P_\phi^{0}, P_\phi^{1}) + 2\sqrt{\sum_{k=1}^K w_k^0 \text{ tr} (\Sigma_k^0)} + 2\sqrt{\sum_{k=1}^K w_k^1 \text{ tr} (\Sigma_k^1)} \\
        & \leq W_1 (P_\phi^{0}, P_\phi^{1}) + 4 \sqrt{\epsilon}
    \end{split}
\end{equation*}
This concludes the proof.
\end{proof}

\begin{corollary}[Sub‑domain version of Theorem~\ref{thm:theorem1}]
\label{cor:sub_A8}
For every subgroup $k$ the bound of Theorem~\ref{thm:theorem1}
applies:
\[
\epsilon_{CF}^{k}(h,\Phi)
   \;\le\;
   {\epsilon_F^{\star}}^{k}(h,\Phi)
\;+\;2B_\Phi\,W_{1}\!\bigl(P_{\Phi,k}^{0},P_{\Phi,k}^{1}\bigr).
\]
Multiplying by $w_k$ and summing over $k$, and using
Lemma~\ref{lem:subgroup_decomp}, yields
\[
\epsilon_{CF}(h,\Phi)
   \;\le\;
   \epsilon_{F}^{\star} (h,\Phi)+2B_\Phi\;
          \Bigl(\sum_{k=1}^{K} w_k
                 W_{1}\!\bigl(P_{\Phi,k}^{0},P_{\Phi,k}^{1}\bigr)
          \Bigr)\;
\]
\end{corollary}
\emph{Remark. } Because clustering is \emph{treatment‑agnostic}, each unit’s subgroup index is determined \emph{solely} by the representation
\(\Phi(x)\), independent of the treatment label \(A\).
Consequently, whenever treatment assignment is randomized or
conditionally ignorable given the representation
(\(A \perp k \mid \Phi(x)\)),
the probability of falling into cluster \(k\) is identical across arms:
\(\Pr(k\!\mid\!A{=}0)=\Pr(k\!\mid\!A{=}1)=w_k\),
so the use of $w_k$,$w_k^0$ and  $w_k^1$ is interchangeable. 


\begin{theorem}[SGA Improves Generalization Bounds, Theorem A.11 in \citep{liu2025understanding}]
Under the following assumptions:
\label{thm:proof_weight_subdomain_distance_stronger_inequation2}
\textbf{A1.} For all $k$, the sub-distributions $P_{\Phi,k}^{0}$ and $P_{\Phi,k}^{1}$ are Gaussian distributions with means $m_k^{0}$ and $m_k^{1}$, and covariances $\Sigma_k^{0}$ and $\Sigma_k^{1}$, respectively. The distance between corresponding sub-distributions is less than or equal to the distance between non-corresponding sub-distributions, i.e., $W_1(P_{\Phi,k}^{0}, P_{\Phi,k}^{1}) \leq W_1(P_{\Phi,k}^{0}, P_{\Phi,k^{\prime}}^{1})$ for $k \neq k^{\prime}$.

\textbf{A2.} There exists a small constant $\epsilon >0$, such that $\underset{1\leq k \leq K}{\max}(\text{tr} (\Sigma_k^{0})) \leq \epsilon$ and $\underset{1\leq k \leq K}{\max}(\text{tr} (\Sigma_k^{1})) \leq \epsilon$. 
Then the following inequalities hold:
\begin{equation*}
\begin{split}
\textstyle\sum_{k=1}^K w_k^{1} W_1(P_{\Phi,k}^{0}, P_{\Phi,k}^{1}) \leq W_1(p_\Phi^{0}, p_\Phi^{1}) + \delta_c,
\end{split}
\end{equation*}
where $\delta_c$ is $4\sqrt{\epsilon}$. 
\end{theorem}

\begin{proof}
Let $\mathbf{w^{0}}\doteq[w_1^{0},\dots,w_K^{0}]$ and $\mathbf{w^{1}}\doteq[w_1^{1},\dots,w_K^{1}]$ belong to $\Delta^{K}$ (the $K-1$ probability simplex). $\Pi(w^{0},w^{1})$ represents the simplex $\Delta^{K \times K}$ with marginals $\mathbf{w^{0}}$ and $\mathbf{w^{1}}$. For any $w \in \Pi(w^{0}, w^{1})$, we can express $w_k^{1} = \sum_{k^{\prime}=1}^K w_{k, k^{\prime}}$. Based on assumption A1, we have:

\begin{equation*}
\begin{split}
\sum_{k=1}^K w_k^{1} W_1(P_{\Phi,k}^{0}, P_{\Phi,k}^{1}) &= \sum_{k=1}^K \sum_{k'=1}^K w_{k, k'} W_1(P_{\Phi,k}^{0}, P_{\Phi,k}^{1}) \\
&\leq \sum_{k=1}^K \sum_{k'=1}^K w_{k, k'} W_1(P_{\Phi,k}^{0}, P_{\Phi,k'}^{1}).
\end{split}
\end{equation*}

Thus, we have (with $MW_1(p_\Phi^{0}, p_\Phi^{1})$ defined in Appendix~\ref{defwassm}):

\begin{equation}
\begin{split}
\sum_{k=1}^K w_k^{1} W_1(P_{\Phi,k}^{0}, P_{\Phi,k}^{1}) &  \leq \min_{w \in \Pi(\mathbf{w}^{0}, \mathbf{w}^{1})} \sum_{k=1}^K \sum_{k'=1}^K w_{k, k'} W_1(P_{\Phi,k}^{0}, P_{\Phi,k'}^{1})\\
& = MW_1(p_\Phi^{0}, p_\Phi^{1}).
\end{split}
\end{equation}

From Theorem~\ref{w1_proof}, we have:
\begin{equation}
MW_1(p_\Phi^{0}, p_\Phi^{1}) \leq W_1(p_\Phi^{0}, p_\Phi^{1}) + 4 \sqrt{\epsilon}.
\end{equation}
Combining the above results:
\begin{equation}
\sum_{k=1}^K w_k^{1} W_1(P_{\Phi,k}^{0}, P_{\Phi,k}^{1}) \leq W_1(p_\Phi^{0}, p_\Phi^{1}) + 4 \sqrt{\epsilon}.
\end{equation}
\end{proof}

\textbf{Justifications of Assumptions in Theorem~\ref{thm:weight_subdomain_distance_stronger}.}





\begin{figure*}[htbp!]
    \centering
    \begin{subfigure}[b]{0.24\textwidth}
        \includegraphics[width=\textwidth]{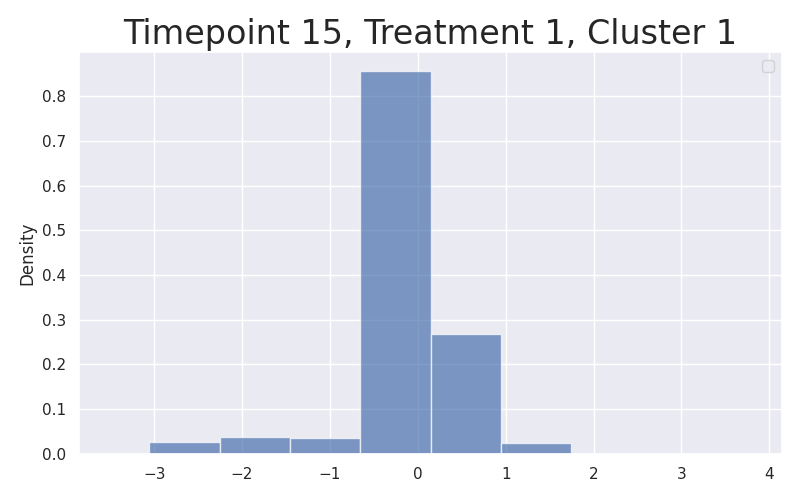} 
        \caption{TP15: Trt 1, Clst 1}
    \end{subfigure}
    \hfill 
    \begin{subfigure}[b]{0.24\textwidth}
        \includegraphics[width=\textwidth]{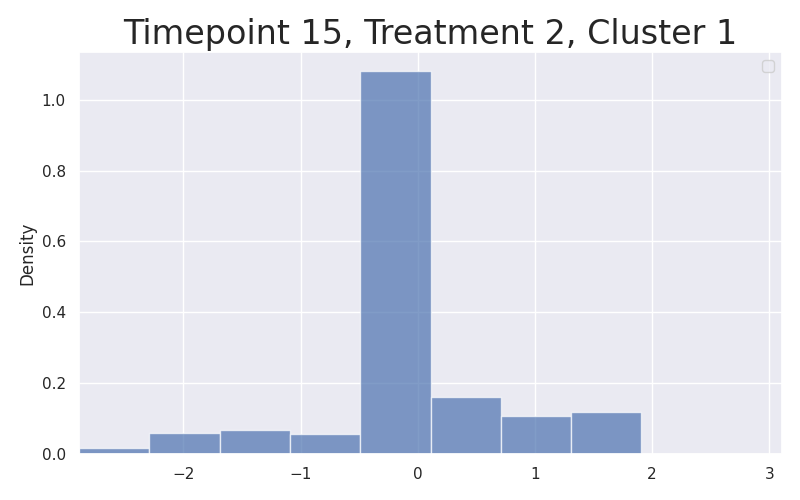} 
        \caption{TP15: Trt 2, Clst 1}
    \end{subfigure}
    \hfill
    \begin{subfigure}[b]{0.24\textwidth}
        \includegraphics[width=\textwidth]{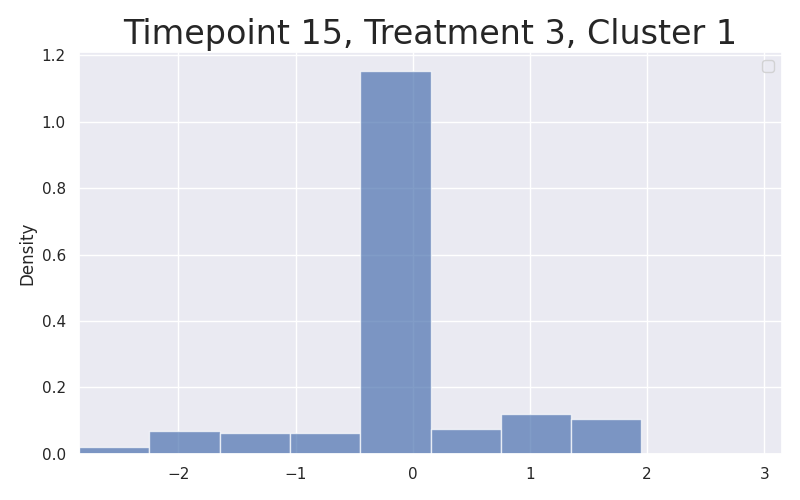} 
        \caption{TP15: Trt 3, Clst 1}
    \end{subfigure}
    \hfill
    \begin{subfigure}[b]{0.24\textwidth}
        \includegraphics[width=\textwidth]{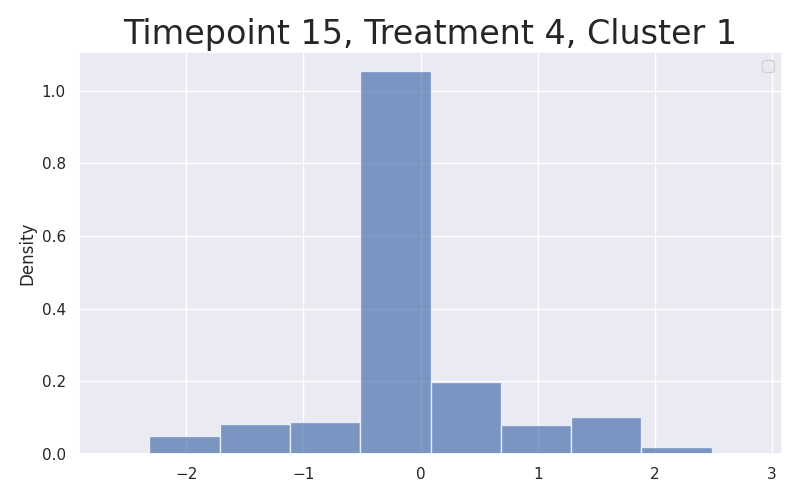} 
        \caption{TP15: Trt 4, Clst 1}
    \end{subfigure}

    \vspace{0.5em} 

    \begin{subfigure}[b]{0.24\textwidth}
        \includegraphics[width=\textwidth]{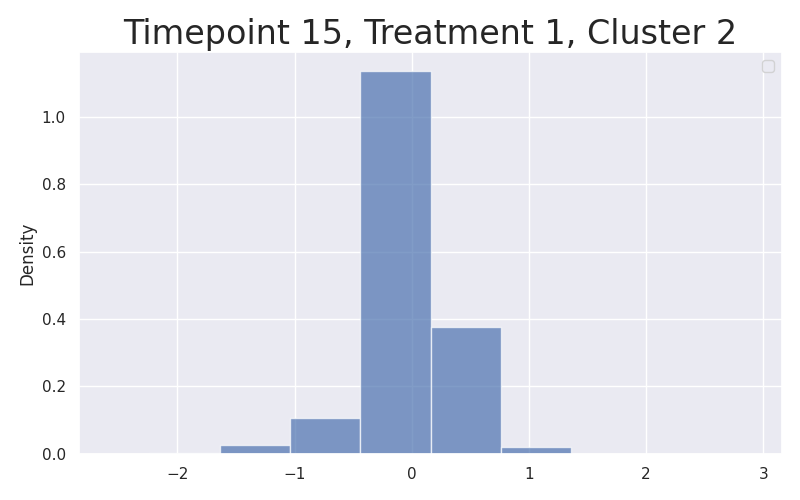} 
        \caption{TP15 Trt 1, Clst 2}
    \end{subfigure}
    \hfill 
    \begin{subfigure}[b]{0.24\textwidth}
        \includegraphics[width=\textwidth]{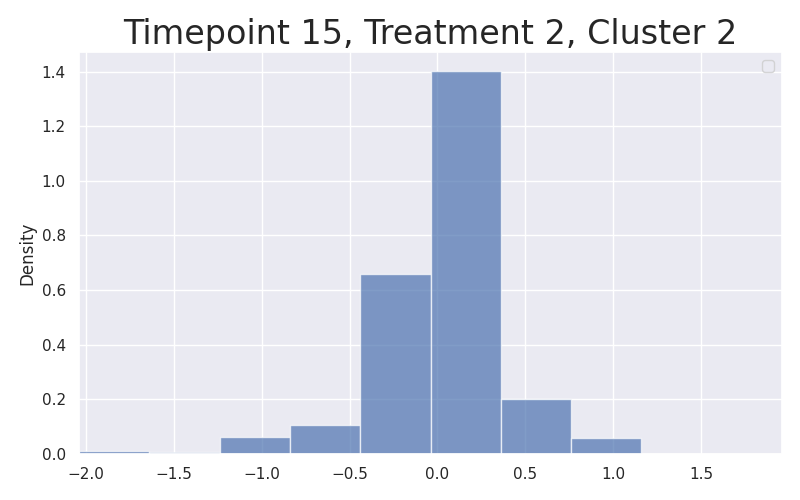}
        \caption{TP15: Trt 2, Clst 2}
    \end{subfigure}
    \hfill
    \begin{subfigure}[b]{0.24\textwidth}
        \includegraphics[width=\textwidth]{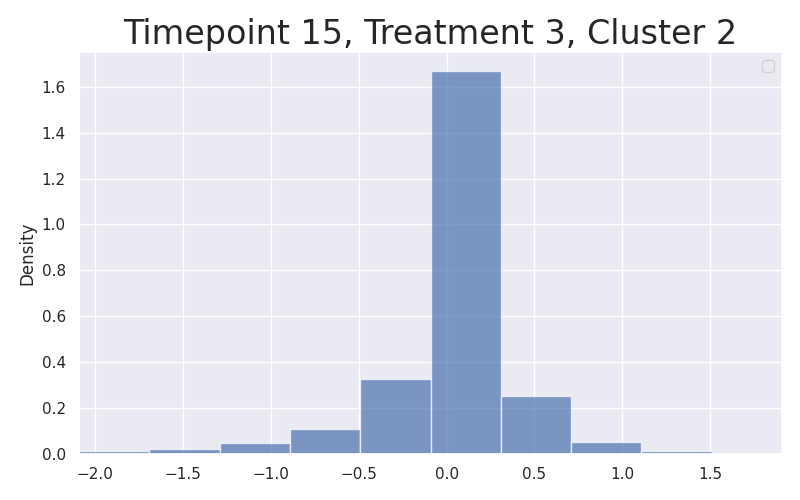}
        \caption{TP15: Trt 3, Clst 2}
    \end{subfigure}
    \hfill
    \begin{subfigure}[b]{0.24\textwidth}
        \includegraphics[width=\textwidth]{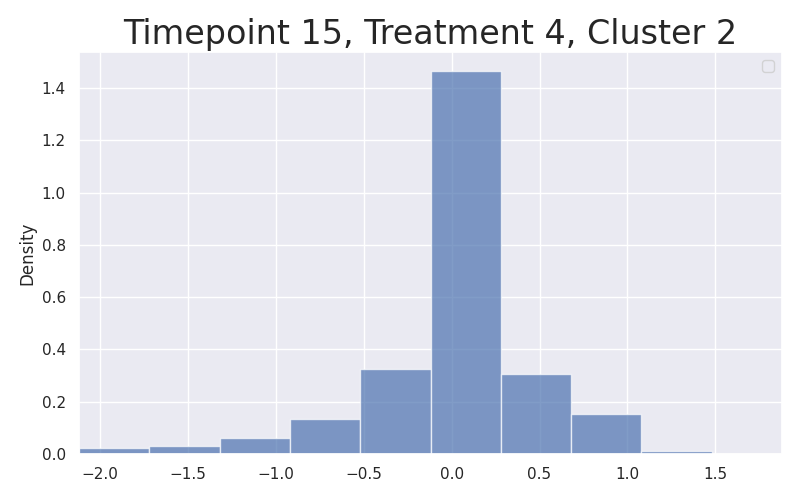}
        \caption{TP15: Trt 4, Clst 2}
    \end{subfigure}

    \vspace{0.5em}

     \begin{subfigure}[b]{0.24\textwidth}
        \includegraphics[width=\textwidth]{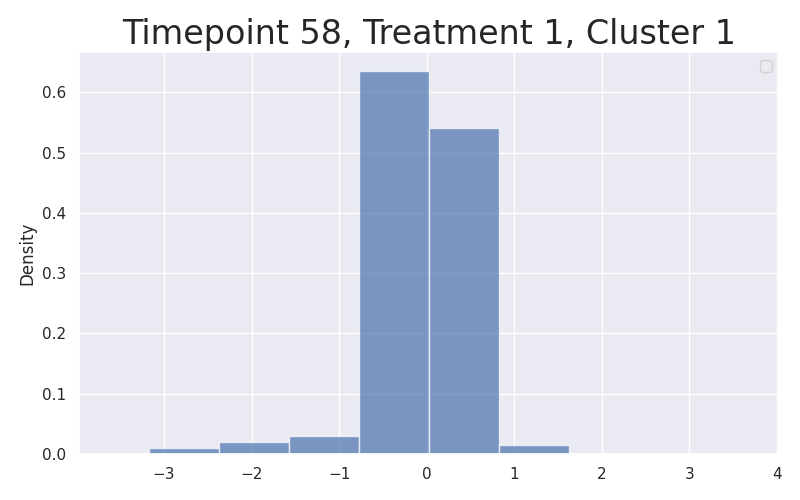} 
        \caption{TP58: Trt 1, Clst 1}
    \end{subfigure}
    \hfill 
    \begin{subfigure}[b]{0.24\textwidth}
        \includegraphics[width=\textwidth]{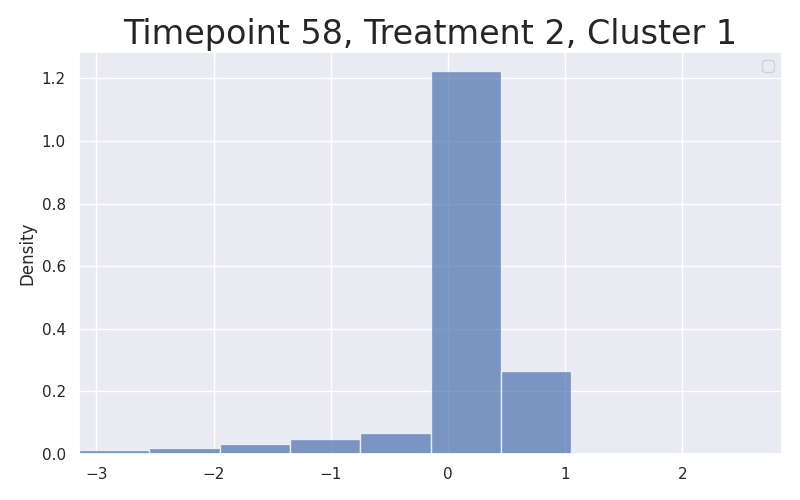} 
        \caption{TP58: Trt 2, Clst 1}
    \end{subfigure}
    \hfill
    \begin{subfigure}[b]{0.24\textwidth}
        \includegraphics[width=\textwidth]{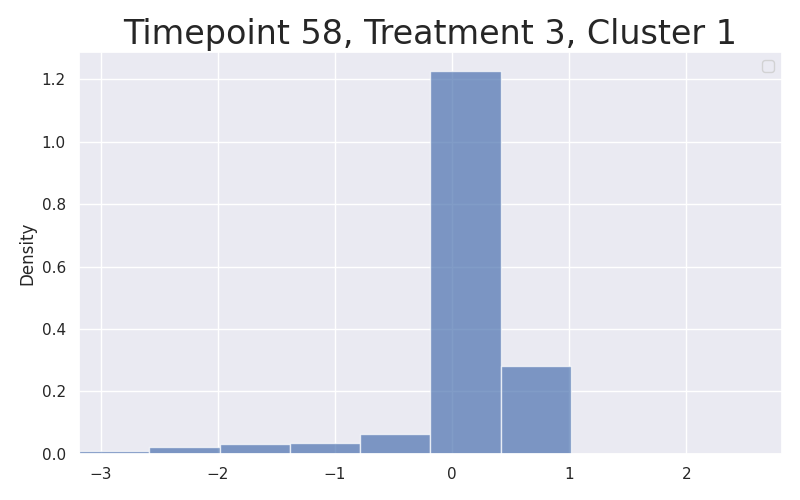} 
        \caption{TP58: Trt 3, Clst 1}
    \end{subfigure}
    \hfill
    \begin{subfigure}[b]{0.24\textwidth}
        \includegraphics[width=\textwidth]{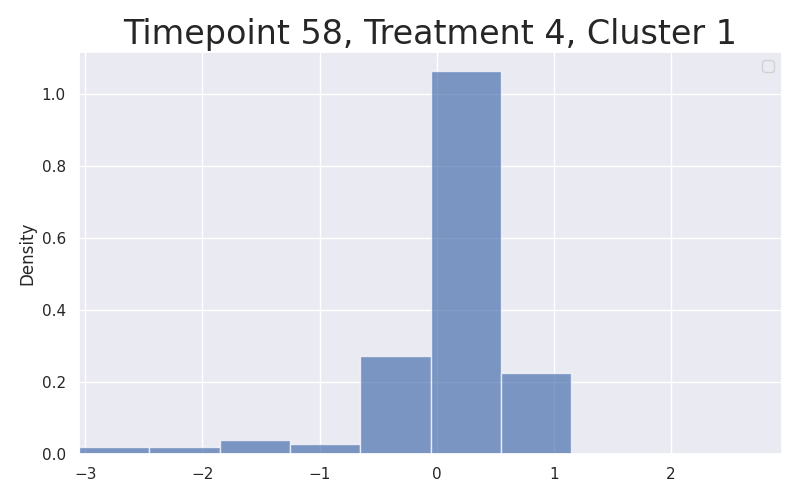} 
        \caption{TP58: Trt 4, Clst 1}
    \end{subfigure}

    \vspace{0.5em} 

    \begin{subfigure}[b]{0.24\textwidth}
        \includegraphics[width=\textwidth]{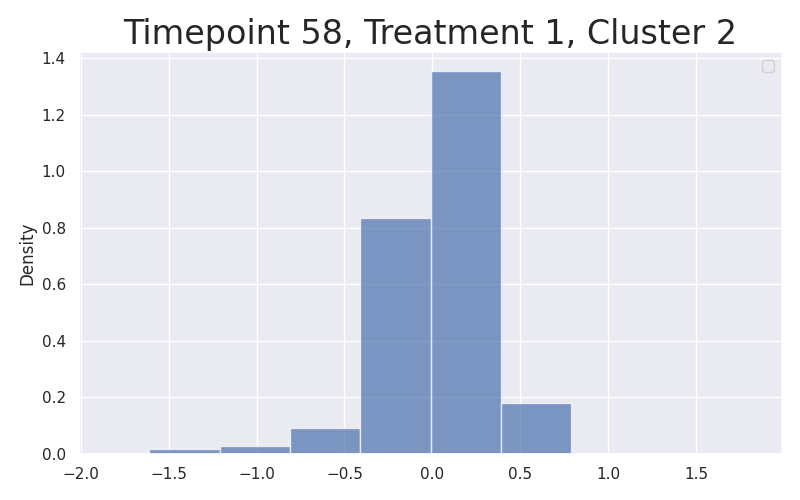} 
        \caption{TP58 Trt 1, Clst 2}
    \end{subfigure}
    \hfill 
    \begin{subfigure}[b]{0.24\textwidth}
        \includegraphics[width=\textwidth]{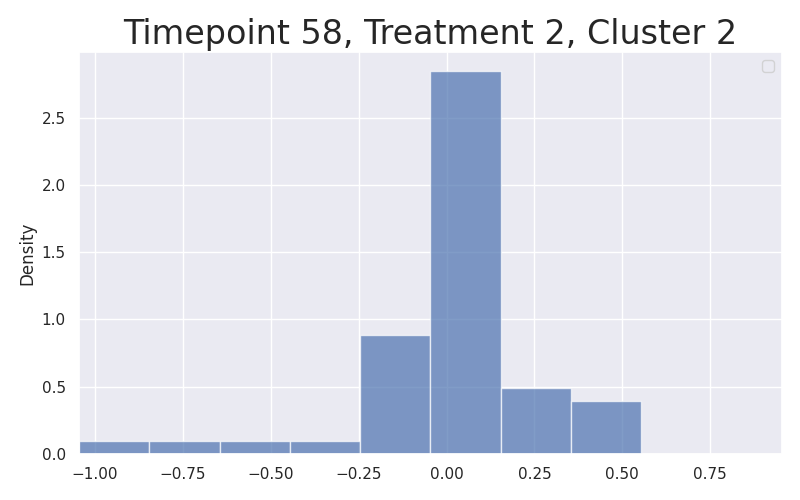}
        \caption{TP58: Trt 2, Clst 2}
    \end{subfigure}
    \hfill
    \begin{subfigure}[b]{0.24\textwidth}
        \includegraphics[width=\textwidth]{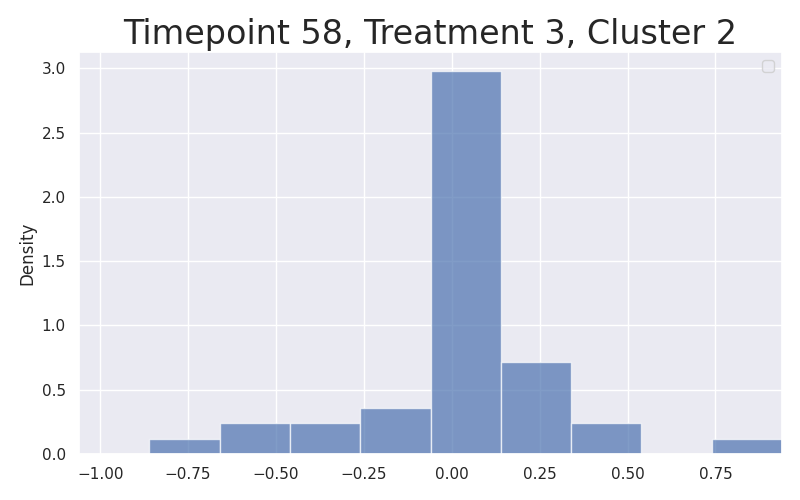}
        \caption{TP58: Trt 3, Clst 2}
    \end{subfigure}
    \hfill
    \begin{subfigure}[b]{0.24\textwidth}
        \includegraphics[width=\textwidth]{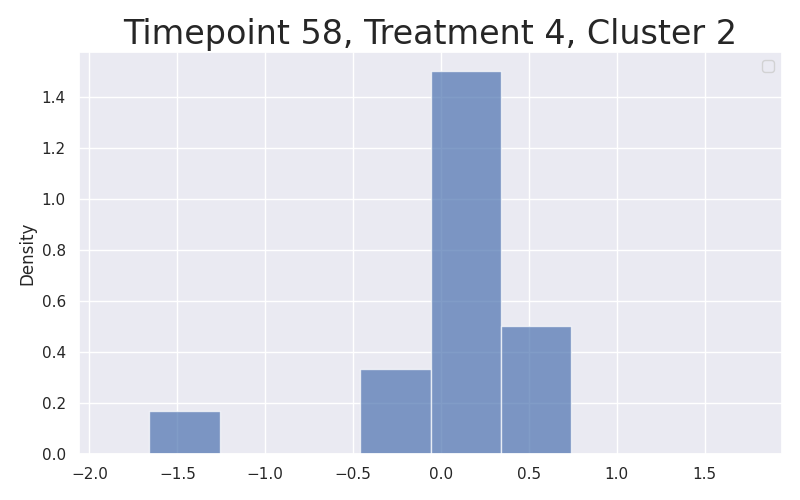}
        \caption{TP58: Trt 4, Clst 2}
    \end{subfigure}
    \caption{\textbf{Distributions of learned representations} for sub-groups within each treatment arm.}
    \label{fig:pca_dist_appendix} 
\end{figure*}

Theorem~\ref{thm:weight_subdomain_distance_stronger} uses the assumptions regarding the distributional properties of sub-treatment groups in the learned representation space $\mathcal{R}$. Specifically, Assumption \textbf{A1} posits that sub-distributions $P_{\Phi,k}^{0}$ and $P_{\Phi,k}^{1}$ (for treatment arms $0, 1$) are Gaussian and that $W_1(P_{\Phi,k}^{0}, P_{\Phi,k}^{1}) \leq W_1(P_{\Phi,k}^{0}, P_{\Phi,k^{\prime}}^{1})$ for $k \neq k^{\prime}$. Assumption \textbf{A2} states that the trace of their covariances is bounded by a small $\epsilon$. We justify the validity of these assumptions below:

\begin{enumerate}[label=\arabic*., leftmargin=*, itemsep=0.5em]
    \item \textbf{Gaussian Sub-distributions in Learned Representation Space (A1, first part):}
    The assumption that sub-treatment groups exhibit Gaussian distributions is made primarily for mathematical tractability in proving the tighter bound. While raw input covariates in complex real-world or semi-synthetic datasets are unlikely to be perfectly Gaussian, this assumption applies to the distributions \emph{after} they have been processed by the encoder network $\phi_E$ into the representation space $\Phi(\mathbf{H}_t)$.
    \begin{itemize}
        \item \textbf{Effect of Deep Encoders:} Deep neural networks, through multiple layers of non-linear transformations, can often map complex, high-dimensional input distributions into lower-dimensional latent spaces where the resulting distributions of distinct sub-group tend to be more regular and sometimes approximate unimodal, bell-shaped forms.
        \item \textbf{Support from Latent Representations Figures:} The histograms in Figure~\ref{fig:pca_dist_appendix} show the learned representations for (treatment, cluster) combinations at selected timepoints. Here we use PCA to map the latent representations to low dimensions. Many of these distributions (Figure~\ref{fig:pca_dist_appendix}), exhibit clear unimodal and symmetric shapes, strongly suggestive of an underlying Gaussian or near-Gaussian structure in that principal component's direction. 
        \item \textbf{Robustness of SGA Implementation:} Importantly, the practical implementation of SGA does not strictly require perfect Gaussianity. Our sensitivity analysis in Section~\ref{sec:clustering_sensitivity_summary} shows that SGA performs well when using k-means for clustering, which imposes fewer explicit distributional assumptions than GMM. This empirical robustness suggests that the benefits of SGA extend beyond scenarios where the Gaussian assumption holds perfectly.
    \end{itemize}

    \item \textbf{Distance Assumption (A1, second part):}
    The assumption $W_1(P_{\Phi,k}^{0}, P_{\Phi,k}^{1}) \leq W_1(P_{\Phi,k}^{0}, P_{\Phi,k^{\prime}}^{1})$ for $k \neq k^{\prime}$ (where $0, 1$ are distinct treatment arms) posits that the Wasserstein-1 distance between \emph{corresponding} sub-groups (same $k$) across two different treatment arms is less than or equal to the distance between a sub-group in one arm and a \emph{non-corresponding} sub-group (different $k'$) in the other arm. 
    
    We provide strong empirical evidence supporting this assumption from our fully-synthetic dataset experiments. For each combination of timepoint, sub-group $k$, and pair of distinct treatment arms $(A, B)$, we calculated the $W_1(P_{\Phi,k}^{A}, P_{\Phi,k}^{B})$ (paired distance) and the average $W_1(P_{\Phi,k}^{A}, P_{\Phi_{k' \neq k}}^{B})$ (average non-paired distance).
    As shown in Table~\ref{tab:distance_assumption_evidence}, the paired Wasserstein distance is less than the average non-Paired Wasserstein distance. Table~\ref{tab:distance_assumption_evidence} presents a detailed view of all such pairwise comparisons for selected representative timepoints (early, mid, and late). This strongly supports Assumption A1 by demonstrating that corresponding sub-groups are indeed closer to each other than to non-corresponding ones in the learned representation space.

    \item \textbf{Bounded Covariance Trace (A2):}
    This assumption ($\underset{1\leq k \leq K}{\max}(\text{tr} (\Sigma_k^{a})) \leq \epsilon$ for any treatment arm $a$) implies that the learned representations of sub-treatment groups are relatively compact. This is a desirable property of representations since well-learned representations for distinct sub-groups are expected to be somewhat concentrated.
\end{enumerate}


\begin{table*}[htbp!]
\centering
\caption{\textbf{Detailed empirical validation of the distance assumption} ($W_1(P_{\Phi,k}^{A}, P_{\Phi,k}^{B}) \leq W_1(P_{\Phi,k}^{A}, P_{\Phi,k^{\prime}}^{B})$ for $k \neq k^{\prime}$) from the fully-synthetic dataset on high confounding $\gamma$=6. The table shows all pairwise treatment comparisons for selected timepoints and sub-groups.}
\label{tab:distance_assumption_evidence}
\resizebox{\textwidth}{!}{%
\begin{tabular}{@{}cccccc@{}}
\toprule
Timepoint & Sub-group $k$ & Treat. Pair (A, B) & Paired $W_1$ & Avg. Non-Paired $W_1$ & Paired $<$ Non-Paired \\
          &               &                    & $(\Phi_k^A, \Phi_k^B)$ & $(\Phi_k^A, \Phi_{k'\neq k}^B)$ & \\ \midrule
\multicolumn{6}{l}{\textit{Timepoint 3}} \\
3 & 1 & (1, 2) & 0.041 & 4.744 & True \\
3 & 1 & (1, 3) & 0.058 & 4.842 & True \\
3 & 1 & (1, 4) & 0.331 & 6.317 & True \\
3 & 1 & (2, 3) & 0.024 & 4.672 & True \\
3 & 1 & (2, 4) & 0.352 & 6.156 & True \\
3 & 1 & (3, 4) & 0.314 & 6.165 & True \\ \addlinespace
3 & 2 & (1, 2) & 0.567 & 3.563 & True \\
3 & 2 & (1, 3) & 0.657 & 3.575 & True \\
3 & 2 & (1, 4) & 2.509 & 4.015 & True \\
3 & 2 & (2, 3) & 0.119 & 4.585 & True \\
3 & 2 & (2, 4) & 1.183 & 5.009 & True \\
3 & 2 & (3, 4) & 1.046 & 5.105 & True \\ \midrule

\multicolumn{6}{l}{\textit{Timepoint 30}} \\
30 & 1 & (1, 2) & 0.092 & 3.565 & True \\
30 & 1 & (1, 3) & 0.099 & 3.458 & True \\
30 & 1 & (1, 4) & 0.448 & 3.466 & True \\
30 & 1 & (2, 3) & 0.011 & 3.124 & True \\
30 & 1 & (2, 4) & 0.178 & 3.132 & True \\
30 & 1 & (3, 4) & 0.170 & 3.119 & True \\ \addlinespace
30 & 2 & (1, 2) & 0.085 & 3.260 & True \\
30 & 2 & (1, 3) & 0.126 & 3.246 & True \\
30 & 2 & (1, 4) & 0.189 & 2.785 & True \\
30 & 2 & (2, 3) & 0.108 & 3.218 & True \\
30 & 2 & (2, 4) & 0.168 & 2.756 & True \\
30 & 2 & (3, 4) & 0.176 & 2.649 & True \\ \midrule

\multicolumn{6}{l}{\textit{Timepoint 59}} \\
59 & 1 & (1, 2) & 0.048 & 3.821 & True \\
59 & 1 & (1, 3) & 0.056 & 3.514 & True \\
59 & 1 & (1, 4) & 0.288 & 3.306 & True \\
59 & 1 & (2, 3) & 0.012 & 3.339 & True \\
59 & 1 & (2, 4) & 0.164 & 3.130 & True \\
59 & 1 & (3, 4) & 0.138 & 3.091 & True \\ \addlinespace
59 & 2 & (1, 2) & 0.170 & 3.447 & True \\
59 & 2 & (1, 3) & 0.113 & 3.407 & True \\
59 & 2 & (1, 4) & 0.310 & 3.037 & True \\
59 & 2 & (2, 3) & 0.233 & 3.608 & True \\
59 & 2 & (2, 4) & 0.531 & 3.238 & True \\
59 & 2 & (3, 4) & 0.269 & 2.929 & True \\ \bottomrule
\end{tabular}%
}
\end{table*}

\section{Implementation Details and Algorithm}
\label{sec:algo}

\paragraph{Training Procedure with Warmup and Epoch-wise SGA}
\label{par:warmupSGA}
In order to stabilize the encoder representations before introducing the Sub-treatment Group Alignment (SGA) objective, we use a \emph{warmup} phase where we optimize a baseline method's losses (e.g., factual outcome prediction plus any other baseline objectives) \emph{without} the SGA term. This is because early in training, the representations encoded by \(\theta_E\) may be very noisy and prone to random alignment. This ensures that the model achieves a coherent representation of the data before the sub-treatment groups are forced to align. Once these representations are sufficiently stable, we then \emph{periodically} compute and backpropagate SGA at specific epochs (e.g., every \(\text{gap\_epoch}\)). This strategy limits the noise in cluster assignments and reduces the computational overhead of repeatedly clustering on each iteration.

\paragraph{Computing the Uniform Mixture of Sub-treatment Groups} In our implementation of the SGA loss, for each time step $t$ and each cluster $k$, we compute the uniform mixture of sub-treatment groups $\phi_E^{t,k}$. \\
To compute this uniform mixture, we perform the following steps: \\
\begin{enumerate}
    \item \textbf{Concatenate representations across treatments:}\\
    For the $k$-th cluster at time $t$, we collect the representations from all treatment groups:
    \[
    \phi_E^{t,k}(\mathbf{H}_t) = \bigcup_{a \in \mathcal{A}} \phi_E^{t,a,k}(\mathbf{H}_t),
    \]
    where $\phi_E^{t,a,k}(\mathbf{H}_t)$ denotes the representations of samples in the $k$-th sub-group of treatment $a$ at time $t$.
    \item \textbf{Shuffle and subsample:}\\
    We shuffle the concatenated representations to ensure that samples from different treatments are thoroughly mixed. Then we select $\frac{1}{|\mathcal{A}|}$ from the concatenated representations as $\phi_E^{t,k}$.
\end{enumerate}

Full details are included in Algorithm \ref{algorithm:SGA_RTM} 

\textbf{Clustering Correspondence \& Clustering Details.} On a high level, we perform a  \emph{treatment-agnostic clustering} in the using the learned representations at each time step. In particular, the clustering is done iteratively and end-to-end together with model (network) training. 

\emph{A Running Example of Clustering Correspondence.} Specifically, at each time step, we run the clustering algorithm (GMM/K-Means) on all the individuals to partition them into $K$ subgroups. Note that here in each subgroup, we may have individuals within different treatment groups because the clustering is treatment agnostic. For example, suppose there are 20 individuals in treatment group 0 and 30 individuals in treatment group 1. In this case, we run the clustering algorithm on all the 50 (20+30) individuals and partition them in K=3 subgroups, with 15,17,18 individuals in each subgroup respectively, The 15 individuals in the first subgroup may be from different treatment groups (e.g., 7 from treatment group 0 and 8 from treatment group 1). 

In this case, these 7 individuals from treatment group 0 and 8 individuals from treatment group 1 are \emph{corresponding sub-treatment groups}, and we have established a \emph{cluster correspondence} between them, as they all belong to the same subgroup identified by the clustering algorithm. 

\textbf{Hyper-parameters.} Regarding the hyperparameters of RTM, we perform hyperparameter optimization for all benchmarks via random grid search with respect to the factual RMSE of the validation set. For reproducibility,we fixed the masking prob as 5\% for all experiments and report the selected hyperparameters in Section~\ref{supp:experiments}. 

Regarding the deep sequence model hyperparameters, to ensure a fair comparison, \emph{we did not tune or re-implement the CT/CRN baseline}. Instead, we directly used the architecture / model hyperparameters reported in the original papers and their official code repositories.

\section{Experiments}
\label{supp:experiments}



\subsection{Fully Synthetic Dataset}
\label{sup:exp_fully_synthetic}

\subsubsection{Dataset Generation}

Dataset generation follows the identical setup as \cite{bica2020estimating, melnychuk2022causal}. The tumor growth simulator \citep{geng2017prediction} models the tumor volume $Y_{t+1}$ after $t+1$ days of diagnosis. It includes two binary treatments: (i) radiotherapy $A_t^r$ and (ii) chemotherapy $A_t^c$. These treatments influence tumor progression as follows:

\begin{itemize}
    \item \textbf{Radiotherapy} has an immediate impact, denoted by $d(t)$, on the tumor volume at the next time step.
    \item \textbf{Chemotherapy} impacts future tumor progression with an exponentially decaying effect $C(t)$.
\end{itemize}

The model is described by the equation:

\[
Y_{t+1} = \left(1 + \rho \log\left(\frac{K}{Y_t}\right) - \beta_c C_t - (\alpha_r d_t + \beta_r d_t^2) + \varepsilon_t \right) Y_t
\]

where $ \varepsilon_t \sim N(0, 0.01^2)$ is independent noise, and the variables $ \beta_c, \alpha_r, \beta_r$ represent the response characteristics for each individual. These parameters are drawn from truncated normal distributions comprising three mixture components. For a full list of parameter values, the code implementation should be consulted.

Time-varying confounding is accounted for through biased treatment assignments, where treatment allocation is identical across both therapies $ A_t^r $ and $ A_t^c $:

\[
A_t^r, A_t^c \sim \text{Bernoulli}\left(\sigma\left(\frac{\gamma}{D_{\text{max}}}(\bar{D}_{15} \bar{Y}_{t-1} - \frac{D_{\text{max}}}{2}) \right)\right)
\]

In this formula, $\sigma(\cdot)$ represents the sigmoid function, $D_{\text{max}}$ is the maximum tumor diameter in the last 15 days, and $\gamma$ is the confounding parameter. $ \bar{D}_{15}(\bar{Y}_{t-1}) $ refers to the average tumor diameter over the previous 15 days. If $\gamma = 0$, the treatment assignment is fully randomized, but for increasing values of $\gamma$, time-varying confounding gradually intensifies. More details can be found in Appendix J in CT \cite{melnychuk2022causal}.

\begin{algorithm}[h]
\caption{Counterfactual Outcome Estimation with Sub-treatment Group Alignment (SGA) and Random Temporal Masking (RTM)} 
\label{algorithm:SGA_RTM}
\begin{algorithmic}[1]
\REQUIRE
\STATE $\mathcal{D} = \{(\mathbf{X}_i^{t}, \mathbf{A}_i^{t}, \mathbf{Y}_i^{t+1})\}_{i=1}^N$: Training data for $N$ individuals for t = 1,...,T
\STATE $\theta_E$, $\theta_Y$: Parameters of encoder $\phi_E$ and regressor $f_Y$
\STATE $\lambda$: Hyperparameter for $L_D$
\STATE $K$: Number of sub-treatment groups (clusters)
\STATE $\mathcal{A}$: Set of possible treatments
\STATE $\text{MaskProb}$: Probability of masking covariates in RTM
\STATE $\eta$: Learning rate
\STATE $\ell(\cdot,\cdot)$: Loss function (e.g., mean squared error)
\STATE $\text{pretrain\_epochs}$: number of epochs to train only $L_Y$ (SGA off)
\STATE $\text{gap\_epoch}$: interval at which SGA is turned on (e.g. 5,10,15,...)
\STATE $\text{max\_epochs}$: total number of training epochs

\STATE  \textbf{Apply Random Temporal Masking (RTM):}
    \FOR{each time step $t = 1$ to $T$}
        \STATE With probability $\text{MaskProb}$, replace $X^t$ with Gaussian noise
    \ENDFOR
\FOR{\text{epoch} = 1 \text{ to } \text{max\_epochs}}
    \STATE Initialize $L_Y = 0, L_D = 0$

    \FOR{each time step $t = 1$ to $T$}
        \STATE $\Phi_E(\mathbf{H}_{t}) = \phi_E(\mathbf{H}_{t}, A^t)$
        \STATE $\hat{Y} = f_Y\left(\Phi_E(\mathbf{H}_{t})\right)$ 
        \STATE \textbf{Compute Factual Outcome Loss:}
        \STATE $L_Y = L_Y + \ell(Y^{t+1}, \hat{Y}^{t+1})$

        \IF{(\text{epoch} $\ge$ \text{pretrain\_epochs}) \textbf{ AND } ((\text{epoch} $\bmod$ \text{gap\_epoch}) = 0)}
            \STATE \textbf{Compute SGA Loss:}
            \FOR{each treatment $a \in \mathcal{A}$}
                \STATE \textbf{Cluster representations into $K$ sub-groups:}
                \STATE Apply GMM to $\Phi_E (\mathbf{H}_{t})$ to obtain clusters $\{\phi_E^{t,a,k}\}_{k=1}^K$
                \STATE Compute weights $w_k^{t,a} = \frac{n_k^{t,a}}{n^{t,a}}$, where $n_k^{t,a}$ is the number of samples in cluster $k$, $n^{t,a}$ is the total number of samples with treatment $a$ at time $t$
            \ENDFOR

            \STATE \textbf{Compute uniform mixture of sub-groups $\phi_E^{t,k}$}
            \STATE \textbf{Compute SGA loss at time $t$:}
            \STATE $L_D = L_D + \sum_{k=1}^K \sum_{a \in \mathcal{A}} w_k^{t,a} \cdot W_1\left( \phi_E^{t,a,k}, \phi_E^{t,k} \right)$
        \ENDIF
    \ENDFOR

    \STATE \textbf{Compute Total Loss:}
    \IF{(\text{epoch} $\ge$ \text{pretrain\_epochs}) \textbf{ AND } ((\text{epoch} $\bmod$ \text{gap\_epoch}) = 0)}
        \STATE $L = L_Y + \lambda L_D$
    \ELSE
        \STATE $L = L_Y$ + \text{baseline suggested loss}
    \ENDIF

    \STATE $\theta_E \leftarrow \theta_E - \eta \nabla_{\theta_E} L$
    \STATE $\theta_Y \leftarrow \theta_Y - \eta \nabla_{\theta_Y} L$

\ENDFOR
\end{algorithmic}
\end{algorithm}

\subsubsection{Experiments Setup}

\textbf{One-step-ahead prediction.} To evaluate one-step-ahead predictions, we utilize the counterfactual trajectories simulated in CT. Our approach involves comparing our estimated outcomes $Y_{t+1}$ against all four possible combinations of one-step-ahead counterfactual outcomes. This effectively captures the tumor volumes under every possible treatment assignment at the next time step.

\textbf{$\tau$-step-ahead prediction.} For multi-step-ahead predictions, the number of potential outcomes for $Y_{t+2}$,...,$Y_{t+\tau_{max}}$ grows exponentially with the prediction horizon $\tau_{max}$.  To manage this complexity, and following the methodology in CT, we employ a single sliding treatment strategy. This approach is motivated by the importance of treatment timing in clinical settings. As discussed in the introduction, consider the treatment of \emph{Ductal Carcinoma In Situ}, where the timing of surgical intervention is critical: delaying surgery might allow the cancer to progress to an invasive stage, while performing it too early could lead to unnecessary invasiveness. To assess whether our models can identify the optimal timing for treatment, we simulate trajectories with a single treatment event that is iteratively shifted across a window ranging from time t to $t + \tau_{max} - 1$. 

\textbf{Performance evaluation.}In line with \citet{melnychuk2022causal}, we evaluate model performance using the mean Root Mean Square Error (RMSE) on the test set, which consists of hold-out data. The RMSE is normalized by dividing by the maximum tumor volume $V_{max} = 1150 \text{cm}^3$. Additionally, we report the test RMSE calculated exclusively on the counterfactual outcomes following the rolling origin, thereby isolating the evaluation from historical factual patient trajectories.

\subsubsection{Empirical Analysis of our Proposed Generalization Bound}
\label{emp_tighter_more}

\begin{figure*}[h]
\centering
\begin{subfigure}{0.44\textwidth}
    \includegraphics[width=\linewidth]{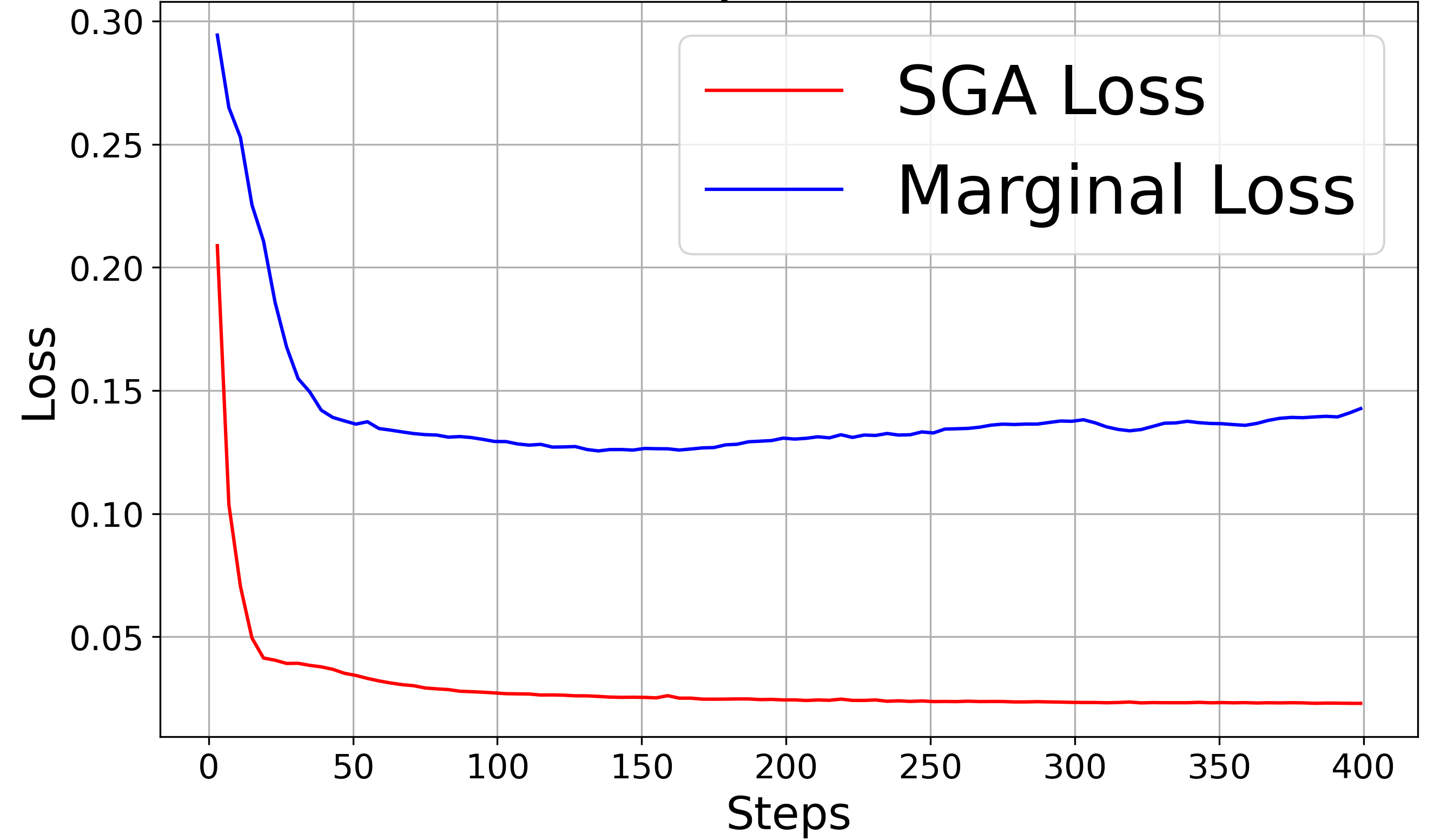}
    \caption{Confounding level - $\gamma$ = 1}
\end{subfigure}
\hspace{2mm}
\begin{subfigure}{0.44\textwidth}
\includegraphics[width=\linewidth]{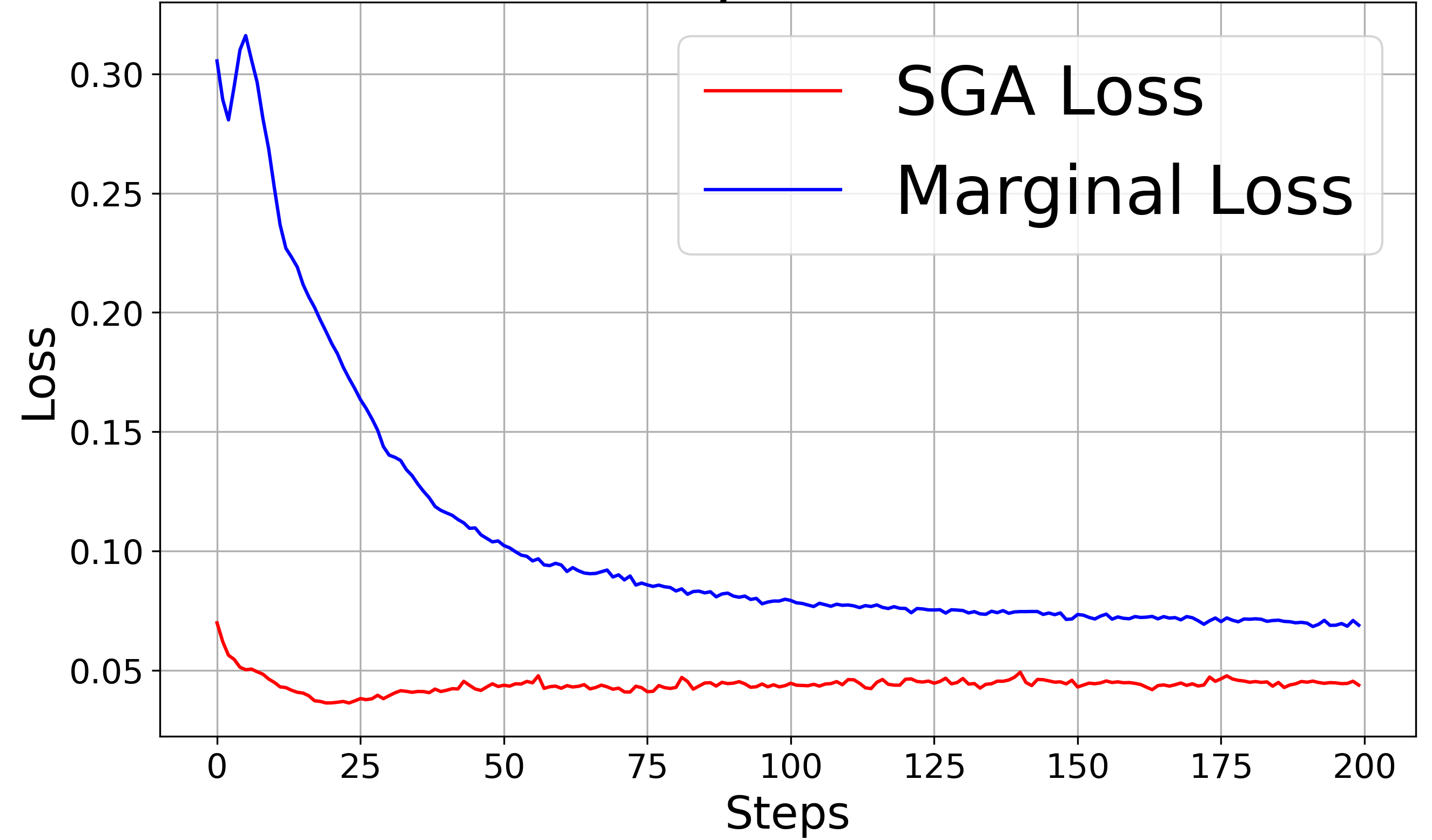}
    \caption{Confounding level - $\gamma$ = 2}
\end{subfigure} \\
\begin{subfigure}{0.44\textwidth}
    \includegraphics[width=\linewidth]{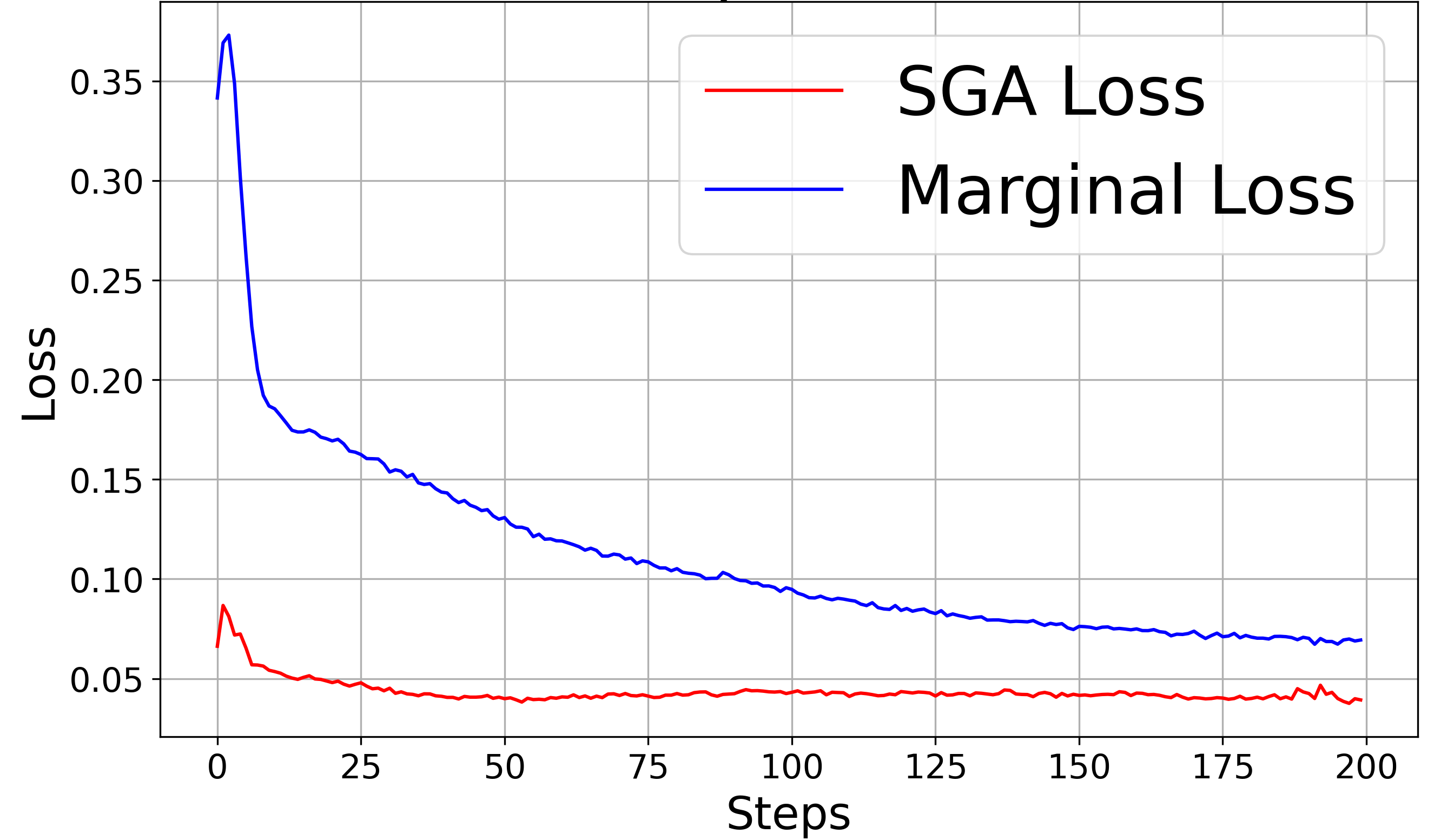}
    \caption{Confounding level - $\gamma$ = 3}
\end{subfigure}
\hspace{2mm}
\begin{subfigure}{0.44\textwidth}
\includegraphics[width=\linewidth]{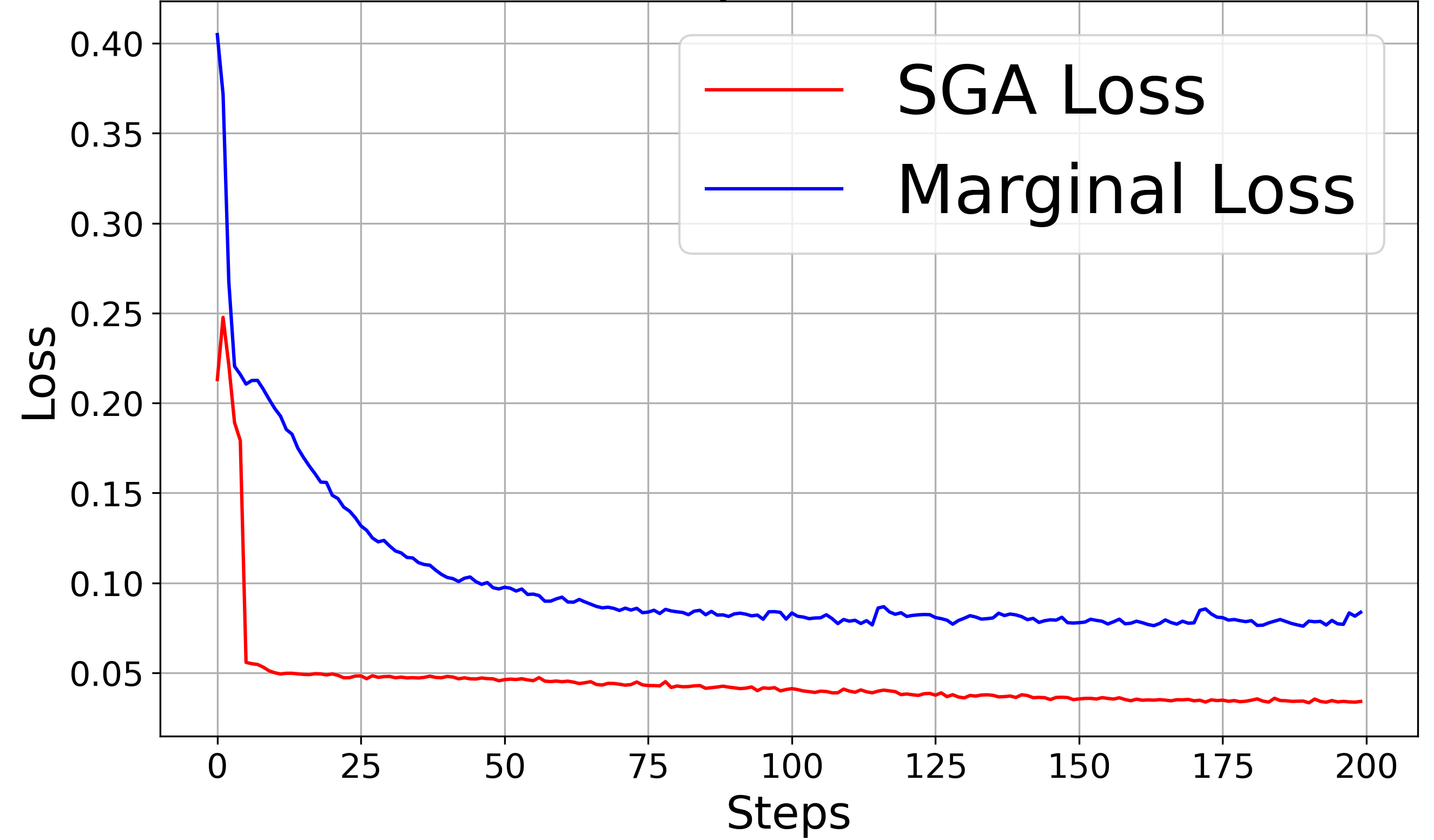}
    \caption{Confounding level - $\gamma$ = 4}
\end{subfigure}
\caption{Empirical results for Sub-treatment Group Alignment (SGA) vs. the original method in Theorem~\ref{thm:theorem1} with varying confounding levels.}
\label{fig:5}
\end{figure*}

As shown in Fig~\ref{fig:5}, here we empirically evaluate the proposed generalization bound.  we provide empirical evidence that Sub-treatment Group Alignment (SGA) results in a much tighter upper bound compared to the original method in Theorem~\ref{thm:theorem1}.

\subsubsection{Analysis of Representation Space}

\begin{figure}[h]
\centering
\includegraphics[width=0.6\textwidth]{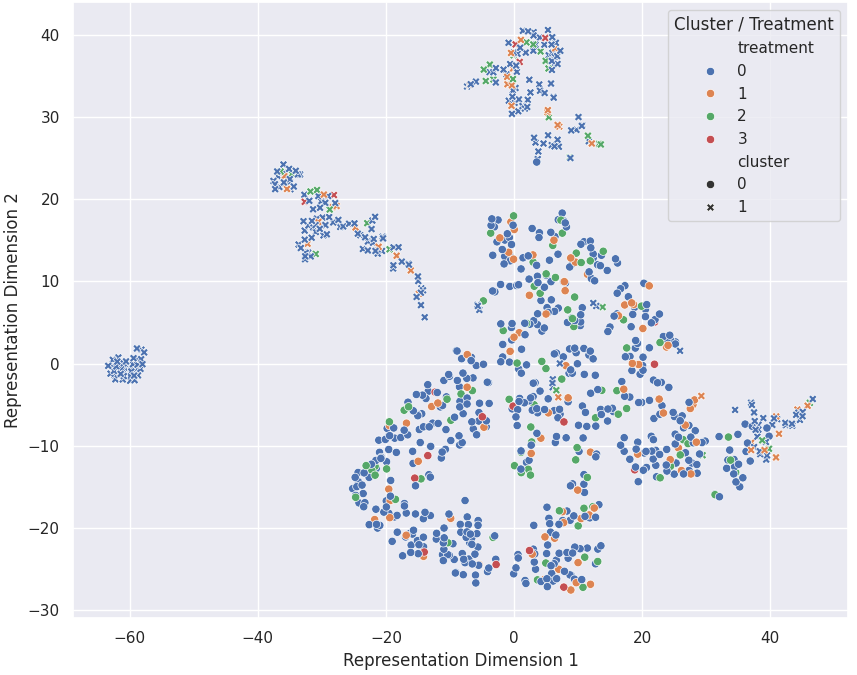}
\caption{\textbf{Representations at the last time point in training under high-confounding scenarios}, with features projected to two dimensions using UMAP.} 
\label{fig:6}
\end{figure}

We visualize the feature spaces learned by our Sub-treatment Group Alignment (SGA) method. As shown in Figure~\ref{fig:6}, SGA is able to learn treatment-invariant representations, which improves performance in counterfactual outcome estimation.

\subsubsection{Analysis of Masking Frequency}

Here, we highlight the potential impact of RTM masking frequency. Indeed, RTM works on a trade-off: by masking covariates, it forces the model to rely more on historical context, aiming to learn more robust causal patterns and reduce overfitting to potentially noisy current-step information. However, if this masking becomes too aggressive, the loss of immediate information could potentially degrade performance on tasks heavily reliant on that information. Experiments with varying masking frequencies with high confounding level $\gamma=6$ on fully synthetic data are shown in Table~\ref{tab:masking_performance_varying_prop}.

\begin{table}[htbp] 
\centering
\caption{Performance metrics for \textbf{different masking frequencies}.}
\label{tab:masking_performance_varying_prop} 
\begin{tabular}{|c|c|c|c|} 
\hline 
Masking Freq & $\tau=4$ & $\tau=5$ & $\tau=6$ \\
\hline 
0\%            &   3.516    &  3.570    &  3.436     \\
2\%            &   2.677    &  2.755     &  2.756     \\
5\%            &   2.009    &  2.039    &  2.005     \\
10\%           &   2.897    &  2.986     &  3.029      \\
20\%           &   3.519    &  3.612    &  3.610      \\
50\%           &   6.630    &  6.863    &  7.006      \\
\hline 
\end{tabular}
\end{table}

As observed, performance initially improves with masking (peaking around 5\%) but degrades significantly with higher masking frequencies, confirming the trade-off.

\subsubsection{Error Bars}
\label{error_bars}

\begin{figure}[h]
    \centering
    \includegraphics[width=\textwidth]{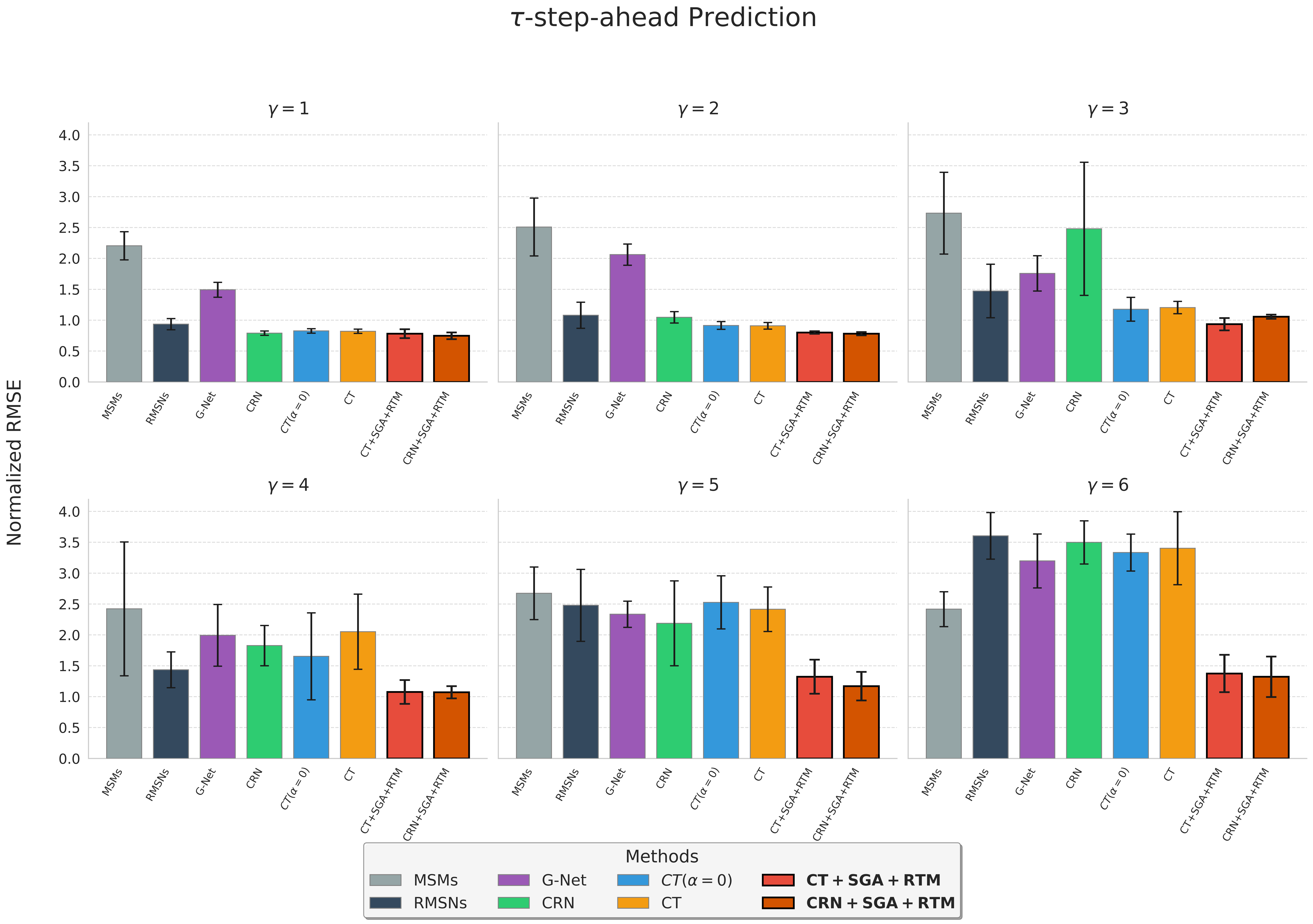} 
    \caption{
        \textbf{Impact of varying confounding level $\gamma$ on long-term ($\tau=6$) prediction performance with error bars}. This figure extends analysis from Figure\ref{fig:3} by incorporating error bars.
    }
    \label{fig:gamma_effect_tau6_appendix}
\end{figure}

\paragraph{Extended Discussion on Sensitivity to varying confounding levels for Long-Term Predictions ($\tau=6$)}
Figure~\ref{fig:gamma_effect_tau6_appendix} shows the robustness of our proposed framework when subjected to varying confounding levels, specifically for the challenging task of long-term counterfactual outcome prediction at $\tau=6$. The Normalized RMSE, along with error bars indicating the spread of results (e.g., mean $\pm$ standard deviation), is reported for our methods (CT+SGA+RTM and CRN+SGA+RTM) and several baselines across six different settings of $\gamma$. The results highlight that both CRN+SGA+RTM and CT+SGA+RTM maintain competitive or superior performance with relatively tight error bars as $\gamma$ changes. This suggests that our synergistic combination of Sub-treatment Group Alignment (SGA) and Random Temporal Masking (RTM) not only achieves high accuracy but also exhibits stability in its predictions for long-range forecasting. The inclusion of error bars provides further confidence in these findings.

\subsubsection{Model Hyperparameters}
Benchmark method hyperparameters and performance are sourced from the GitHub repository of \citet{melnychuk2022causal}.

\begin{table}[H]
 \centering
 \caption{Model hyperparameters used for the fully-synthetic dataset}
 \scalebox{0.8}{
 \begin{tabular}{|c|c|c|} 
 \hline
 ~  &  CT + SGA + RTM  & CRN + SGA + RTM\\ 
 \hline
$\gamma=0$  & \begin{tabular}[c]{@{}c@{}} batch size = 2048, lr = 0.025, \\ $\lambda$ = 0.0001, \\ dropout rate = 0.2,  Adam~\end{tabular}  & \begin{tabular}[c]{@{}c@{}} encoder batch size = 1024, encoder lr = 0.005, encoder dropout rate = 0.1, \\ decoder batch size = 4096, decoder lr = 0.01, decoder dropout rate = 0.2, \\$\lambda$ = 0.0001, Adam~\end{tabular}  \\
\hline
$\gamma=1$ & \begin{tabular}[c]{@{}c@{}} batch size = 1024, lr = 0.02, \\ $\lambda$ = 0.0001, dropout rate = 0.1, \\ Adam~\end{tabular} & \begin{tabular}[c]{@{}c@{}} encoder batch size = 1024,  encoder lr = 0.005, encoder dropout rate = 0.1, \\ decoder batch size = 4096, decoder lr = 0.01, decoder dropout rate = 0.1,\\ $\lambda$ = 0.0001, Adam~\end{tabular}  \\
\hline
$\gamma=2$ & \begin{tabular}[c]{@{}c@{}} batch size = 512, lr = 0.02, \\ $\lambda$ = 0.001, dropout rate = 0.1, \\ Adam~\end{tabular} &  \begin{tabular}[c]{@{}c@{}} encoder batch size = 1024, encoder lr = 0.005, encoder dropout rate = 0.2, \\ decoder batch size = 4096, decoder lr = 0.01, decoder dropout rate = 0.1, \\ $\lambda$ = 0.0001, Adam~\end{tabular} \\
\hline
$\gamma=3$  & \begin{tabular}[c]{@{}c@{}} batch size = 512, lr = 0.03, \\ $\lambda$ = 0.001, dropout rate = 0.1, \\  Adam~\end{tabular} & \begin{tabular}[c]{@{}c@{}} encoder batch size = 1024, encoder lr = 0.005, encoder dropout rate = 0.2, \\ decoder batch size = 4096, decoder lr = 0.01, decoder dropout rate = 0.1 \\ , $\lambda$ = 0.001, Adam~\end{tabular} \\
\hline
$\gamma=4$  & \begin{tabular}[c]{@{}c@{}} batch size = 1024, lr = 0.01, \\ $\lambda$ = 0.001, dropout rate = 0.1, \\ Adam~\end{tabular} & \begin{tabular}[c]{@{}c@{}} encoder batch size = 1024, encoder lr = 0.005, encoder dropout rate = 0.2, \\ decoder batch size = 4096, decoder lr = 0.01, decoder dropout rate = 0.1, \\ $\lambda$ = 0.01, Adam~\end{tabular}  \\
\hline
$\gamma=5$ & \begin{tabular}[c]{@{}c@{}} batch size = 256, lr = 0.01, \\ $\lambda$ = 0.1, dropout rate = 0.1, Adam~\end{tabular} & \begin{tabular}[c]{@{}c@{}} encoder batch size = 64, encoder lr = 0.001, encoder dropout rate = 0.2, \\ decoder batch size = 1024, decoder lr = 0.001, decoder dropout rate = 0.1, \\ $\lambda$ = 0.001, Adam~\end{tabular} \\
\hline
$\gamma=6$  & \begin{tabular}[c]{@{}c@{}} batch size = 256,  lr = 0.005, \\ $\lambda$ = 0.1,  dropout rate = 0.1, \\ Adam~\end{tabular}  & \begin{tabular}[c]{@{}c@{}} encoder batch size = 64, encoder lr = 0.001, encoder dropout rate = 0.2, \\ decoder batch size = 1024, decoder lr = 0.001, decoder dropout rate = 0.1, \\ $\lambda$ = 0.01, Adam~\end{tabular}  \\
\hline

\end{tabular}
}
\end{table}

\subsection{Semi-synthetic Dataset}
\label{sup:exp_semi_synthetic}
We used the identical semi-synthetic dataset generated by \citet{melnychuk2022causal}, which is based on real-world medical data from intensive care units, to validate our model with high-dimensional, long-range patient trajectories. As outlined in \citet{melnychuk2022causal}, this dataset builds on the MIMIC-III dataset and simulates patient trajectories with both endogenous and exogenous dependencies, taking treatment effects into account \citep{johnson2016mimic}. This setup allows us to control for confounding in our experiments. The use of semi-synthetic data is important here, as real-world data lacks ground-truth counterfactuals, which are necessary for evaluating our methods’ performances. To make our manuscript self-sustained, we hereby summarize the setup elaborated in Causal Transformer \citep{melnychuk2022causal}. Full details on the data generation process can be found in Appendix K \citet{melnychuk2022causal}.

Following \citep{melnychuk2022causal}, we utilized MIMIC-extract \citep{wang2020mimic} based on the MIMIC-III dataset \citep{johnson2016mimic}. The data were preprocessed with forward and backward imputation for missing values and standardization of continuous features. Our dataset included 25 time-varying signals and 3 static covariates (gender, ethnicity, age), yielding 44 total features ($d_w = 44$) after one-hot-encoding.


The simulation follows four main steps:
\begin{enumerate}
    \item \textbf{Cohort Selection} \\
    1,000 patients whose ICU stays lasted between 20 and 100 hours are sampled .

    \item\textbf{Untreated Outcomes} \\
    For each patient $i$, simulated $d_y$ untreated outcomes $\mathbf{Z}_t^{j,(i)}$ are simulated by combining:
\begin{itemize}
    \item A B-spline term as an endogenous component 
    \item Random function $g^{j,(i)}(t)$
    \item Exogenous covariate dependencies $f_Z^j(\mathbf{X_t}^{(i)})$
    \item Independent Gaussian noise $\epsilon_t \sim N(0, 0.005^2)$
\end{itemize}

\[
\mathbf{Z}_t^{j,(i)} = \alpha_S^j \text{B-spline}(t) + \alpha_g^j g^{j,(i)}(t) + \alpha_f^j f_Z^j(\mathbf{X_t}^{(i)}) + \epsilon_t
\]

\item \textbf{Treatment Assignment} \\
We generated binary treatment indicators $\mathbf{A_t}^l$, l = $1,...,d_a$, based on previous outcomes and covariates, using a sigmoid function:

\[
p_{\mathbf{A_t}^l} = \sigma(\gamma_A^l \bar{A}_{T_l}(\bar{Y}_{t-1}) + \gamma_X^l f_Y^l(X_t) + b_l)
\]
\[
\mathbf{A_t}^l \sim \text{Bernoulli}(p_{\mathbf{A_t}^l})
\]
Confounding is added by a subset of current time-varying covariates via a random function $f_Y^l(X_t)$, and $f_Y^l(\cdot)$ is sampled from an RFF approximation of a Gaussian process. \\

\item \textbf{Treatment Effects} \\
In this step, treatments are applied to the initial untreated outcomes. We start by setting $\mathbf{Y_1} = \mathbf{Z_1}$, where each treatment $l$ influences an outcome $j$ with an immediate, maximum effect $\beta_{lj}$ after application. The treatment effect occurs within a time window from $t - w^l$ to $t$, with effect decreasing according to an inverse-square decay over time. The effect is also scaled by the treatment probability $p_{\mathbf{A_t^l}}$. When multiple treatments are involved, their combined effect is calculated by taking the minimum across all treatment impacts.

The aggregated treatment effect is given by:

\[
E^j(t) = \sum_{i=t-w^l}^{t} \frac{\min_{l=1,\dots,d_a} \mathbbm{1}_[\mathbf{A_i^l}=1] p_{\mathbf{A_i^l}} \beta_{lj}}{(w^l - i)^2}
\]

\item \textbf{Combining Treatment Effects} \\
We then add the simulated treatment effect $E^j(t)$ to the untreated outcome $Z_t^j$ to get the final outcome:

\[
Y_t^j = Z_t^j + E^j(t)
\]

\item \textbf{Dataset Generation} \\
The semi-synthetic dataset was generated using the above framework. For the exact parameter values used in the simulation, please refer to the GitHub repository of \citet{melnychuk2022causal}. Following the setup in CT, we used the simulated three synthetic binary treatments ($d_a = 3$) and two synthetic outcomes ($d_y = 2$). We also use the identical setup and split the 1000-patient cohort into training, validation, and test sets, with a 60\%/20\%/20\% split. For one-step-ahead prediction, all $2^3 = 8$ counterfactual outcomes were simulated. For multiple-step-ahead prediction, we sampled 10 random trajectories for each patient and time step, with $\tau_{\text{max}} = 10$.
\end{enumerate}

\subsubsection{Model Hyperparameters}

Benchmark method hyperparameters and performance are sourced from the GitHub repository of \citet{melnychuk2022causal}.

\begin{table}[H]
 \centering
 \caption{Model hyperparameters used for the semi-synthetic dataset.}
 \scalebox{0.7}{
 \begin{tabular}{|c|c|} 
 \hline
 CT + SGA + RTM & \begin{tabular}[c]{@{}c@{}} batch size = 64, \\ lr = 0.01, \\ $\lambda$ = 0.0001, \\ dropout rate = 0.1, \\ Adam~\end{tabular}  \\
 \hline
CRN + SGA + RTM & \begin{tabular}[c]{@{}c@{}} encoder batch size = 128, \\ encoder lr = 0.001, \\ encoder dropout rate = 0.1, \\ decoder batch size = 512, \\ decoder lr = 0.0001, \\ decoder dropout rate = 0.1\\, $\lambda$ = 0.0001, \\ Adam~\end{tabular} \\
  \hline
 \end{tabular}
}
\end{table}

\section{Computational Analysis of SGA}
\label{sec:cost}

While we focus on the theoretical analysis using Wasserstein distance, we recognize the necessity of computational studies due to its high computational cost. In particular, when the probability measures have at most n supports, the computational complexity of the Wasserstein distance is on the order of$ O(n^3 log(n))$ \citep{pele2009fast}. In high-dimensional settings (e.g., images with many pixels), the size of these supports can become very large, compounding the computational burden.

To this end, we employ three approaches to improve the computational efficiency: (i) an embedding network to reduce the dimension of the input and the sinkhorn algorithm \citep{altschuler2017near} for estimating the Wasserstein distance. Through an embedding/representation learning network, high-dimensional inputs can be mapped into low-dimensional spaces, thus considerably reducing the computational cost.
(ii) Sinkhorn algorithm for practicality: We use the Sinkhorn algorithm \citep{altschuler2017near} to approximate the Wasserstein distance, as it provides a more tractable approach than directly solving the optimal transport problem. Although this introduces an approximation error, the regularization parameter $\epsilon$ allows us to balance accuracy and efficiency. Notably, the entropic version can achieve an approximate Wasserstein distance computation in $O(n^2)$ \citep{nguyen2022hierarchical} (up to polynomial orders of approximation errors). (iii) Clustering is performed periodically, not every iteration, balancing performance with computational cost and stability (Details discussed in Appendix~\ref{sec:algo}).

\section{Discussions and Limitations}
\label{sec:limitations}

We address the critical challenge of counterfactual outcome estimation in time series by introducing two novel, synergistic approaches: Sub-treatment Group Alignment (SGA) and Random Temporal Masking (RTM). SGA tackles time-varying confounding at each time point by aligning fine-grained sub-treatment group distributions, leading to tighter counterfactual error bound and more effective deconfounding. Complementarily, RTM promotes temporal generalization and robust learning from historical patterns by randomly masking covariates, encouraging the model to preserve underlying causal relationships across time steps and rely less on potentially noisy contemporaneous time steps.

Our comprehensive experiments demonstrate that SGA and RTM are broadly applicable and significantly enhance existing state-of-the-art methods. While each approach individually improves performance, their synergistic combination consistently achieves SOTA performance on both synthetic and semi-synthetic benchmark datasets. Together, SGA and RTM offer a flexible and effective framework for improving causal inference from observational time series data.

In this work, we focus on synthetic and semi-synthetic datasets to enable controlled evaluation of counterfactual outcomes, where ground truth is available, which is a crucial factor for validating methodological performance. Moving forward, applying our method to real-world data is an important next step. In particular, we are actively exploring a case study on interventions in depressive phenotypes in animal models (e.g., varying treatment response depending on phenotype). This experimental setting allows regularly collected behavioral and physiological data, motivating the techniques proposed here. We believe this setting is a natural fit for our framework, and will facilitate real-world validation while maintaining fine-grained observational control.  We then plan to work with clinical observational studies to further apply the work.

\end{document}